\documentclass[11pt]{article} 
\usepackage[T1]{fontenc}
\usepackage{graphicx}
\usepackage{amsmath}
\usepackage{amsfonts}
\usepackage{authblk}
\usepackage{subcaption}  %For2D_pi_vpi_EpsGreedy subfigures
\usepackage{caption}
\usepackage{float}
\usepackage{booktabs}
\usepackage{color,soul}
\usepackage{array}
\usepackage{microtype} % Better line breaking: tighter, fewer overfull/orphan lines
\usepackage{enumitem}  % Compact list spacing,
\usepackage{orcidlink}
\usepackage{authblk}

\usepackage{listings}
\usepackage{xcolor}
\usepackage{booktabs}
\usepackage{caption}
\usepackage{hyperref}
\usepackage{parskip}

\setlist{topsep=2pt,itemsep=1pt,parsep=0pt}

\let\oldthebibliography\thebibliography
\renewcommand{\thebibliography}[1]{%
  \oldthebibliography{#1}%
  \setlength{\itemsep}{0pt plus 0.3pt}%
  \setlength{\parsep}{0pt}%
}

\begin{document}
\title{Offline Policy Evaluation as a decision‑support tool for designing Adaptive Experiments}
%\titlerunning{OPE as a decision‑support tool}
%\title{Offline Policy Learning and Evaluation: benchmarking the design of Adaptive Experiments}

\author[]{João Victor Ferreira Alves \textsuperscript{1 $\star$} \orcidlink{0009-0000-2023-2931}} \author[]{Eduardo Rocha Laurentino \textsuperscript{2 $\star$} \orcidlink{0000-0001-5100-5029}} \author[]{Gustavo de Oliveira Kanno \textsuperscript{3} \orcidlink{0009-0008-7329-3031}} \author[]{Thiago Costa Rizuti da Rocha \textsuperscript{4 $\dagger$} \orcidlink{0009-0005-9708-2253} } \affil[]{Instituto de Ciência e Tecnologia Itaú (ICTi), Brasil }

%
%\email{joaovfalv@gmail.com}}

%\author{Anonymous Authors}
%\authorrunning{Anonymous Authors}
%\institute{Anonymous Institution\\
%\email{anonymous@example.com}}

%
\renewcommand{\thefootnote}{}
\footnotetext{\hspace{-.7cm} 1 joaovfalv@gmail.com 
\\2 eduardo.laurentino@itau-unibanco.com.br 
\\3 gustavo.kanno@itau-unibanco.com.br 
\\4 thiago.rizuti-rocha@itau-unibanco.com.br 
\\$\star$ Equal contribution as first authors. 
\\ $\dagger$Corresponding author.}

\maketitle

\begin{abstract}

We investigate how historical data from fixed randomized experiments (A/B tests) can be used to inform the deployment of adaptive experiments based on contextual bandits. Given data collected under a static allocation, our goal is to assess which adaptive policies, if any, would have outperformed the original design and under what conditions. To this end, we combine off-policy evaluation (OPE) with a controlled warm-start simulation. From logged A/B test data exhibiting heterogeneous treatment effects, we estimate nuisance components and use doubly robust estimators to rank a portfolio of pre-specified adaptive and non-adaptive policies. When ground truth is available, we then deploy the same offline-trained policies in a simulator that reuses the exact data-generating reward probabilities, providing a safe, ground-truth-anchored environment to study the offline-to-online transition under warm starting. Using synthetic randomized controlled trials with known heterogeneity structures and an oracle policy, our results indicate that adaptive, context-aware policies improve upon fixed allocations when meaningful heterogeneity is present, while providing little benefit in its absence. We reinforce our findings on standard open benchmarks (Hillstrom, Criteo Uplift, and LaLonde), reinterpreted through a policy-value and regret perspective. Overall, our results provide a practical methodology for deciding when adaptive experimentation is worth deploying and how to select among competing adaptive policies using existing A/B test data.

\end{abstract}

\section{Introduction}

Randomized controlled trials (RCTs), commonly known as A/B tests, remain the gold standard for causal inference in technology-based companies, clinical research, and social sciences. In a typical A/B test, experimental units are assigned to one of $K$ treatment arms with fixed and uniform probabilities, and the experimenter collects outcome data until a pre-specified sample size is reached. This simplicity is both the method's greatest strength (since it yields unbiased estimates under minimal assumptions) and its most consequential limitation: by treating all individuals identically, A/B tests forgo the opportunity to personalize treatment assignment based on observable characteristics that may moderate treatment effects.

Adaptive experimental designs, Contextual Multi Armed Bandits (CMAB) in particular, address this limitation by sequentially adjusting the assignment policy as data accumulates. At each round $t$, a contextual bandit algorithm observes a feature (context) vector $X_t$, selects an action (treatment) $A_t$ according to a policy $\pi_t$ that depends on the history of past observations, and receives a stochastic reward (outcome) $Y_t$. This adaptivity offers two potential advantages over static designs: it can improve outcomes for participants during the experiment itself, by routing individuals toward more promising treatments; and it can accelerate the identification of heterogeneous treatment effects (HTE), by concentrating exploration where uncertainty is high. These properties have motivated increasing adoption of bandit-based experimentation in domains ranging from digital advertising to clinical trial design and recommendation systems \cite{HadadEtAl2021AdaptiveExperiments}.

However, the transition from A/B testing to adaptive experimentation is rarely straightforward. Deploying a contextual bandit requires confidence that it will outperform the existing static design, yet this confidence is difficult to obtain without running the adaptive experiment itself. Moreover, even when the decision to adopt a contextual bandit is made, the initial deployment phase can still suffer from a cold-start problem: when data are scarce, the algorithm must explore in order to learn which actions are effective for which contexts, and this exploration can reduce short-term performance before sufficient evidence accumulates. \cite{li_contextual_news_2010,bietti_bakeoff_2021}

In this paper, we argue that historical data from completed A/B tests can be systematically leveraged to address the pre-deployment uncertainty problem of adaptive experimentation. Our goal is to use data from fixed A/B tests to anticipate, before deployment, whether adaptive (contextual bandit) policies would improve outcomes relative to the original static design. We propose a two-level framework that transforms static experimental data into actionable intelligence for adaptive experimentation:
\textbf{Level 1 - Off-Policy Evaluation (OPE):} using historical A/B test logs to counterfactually rank candidate adaptive policies before any live deployment; \textbf{Level 2 - Controlled Warm Simulator:} deploying the OPE-selected policies in a simulator whose reward probabilities are exactly known from the synthetic data-generating process, providing a safe, ground-truth-anchored environment in which policies evaluations can be verified online.

\section{Background}

A principled response to this challenge is \emph{off-policy evaluation} (OPE): the problem of estimating the value $V(\pi)$ of a target policy $\pi$ (the expected reward under $\pi$) using data collected by a different, fixed logging policy $\pi_0$, without deploying $\pi$ in the real world \cite{dudik_DR}. In practice, OPE combines nuisance estimation (propensity scores $\hat b(a\mid x)$ and outcome models $\hat\mu_a(x)$) with counterfactual value estimators such as Inverse Propensity Scoring (IPS), Self-Normalized IPS (SNIPS), and especially Doubly Robust (DR) estimators, which are attractive in finite samples because they remain consistent when either the outcome model or the propensity model is correctly specified and can be implemented with sample splitting to reduce overfitting bias \cite{dudik_DR,SNIPS,wang_switch}. In this paper, we restrict our attention to the contextual bandit setting rather than the full sequential reinforcement learning regime.

Among standard OPE estimators, Doubly Robust (DR) is particularly attractive because it combines the strengths of direct modeling and importance weighting. The direct method (DM) estimates the expected outcome under each action through nuisance models $\hat\mu_a(x)\approx \hat{\mathbb{E}}[Y\mid X=x,A=a]$, which can yield low-variance policy-value estimates but may be biased if the outcome model is misspecified. Inverse Propensity Scoring (IPS), in contrast, reweights observed rewards by the ratio $\pi(a\mid x)/\hat b(a\mid x)$ and is unbiased when the logging policy is correctly known, but it can have high variance when propensities are small or when the target policy differs substantially from the behavior policy. DR combines both components: it uses the outcome model as a baseline prediction and adds a bias-correction term based on inverse propensity weighting. As a result, the estimator remains consistent if either the outcome model or the propensity model is correctly specified, which makes it a natural default in finite samples and a strong choice for practical policy screening. In our setting, where the logging policy is fixed and known in the synthetic experiments, DR offers a favorable bias-variance trade-off while retaining a clear counterfactual interpretation \cite{dudik_DR,SNIPS,wang_switch,su2020dros,KallusUehara2020DRL}.

For deployment, one may also consider Offline Policy Learning (OPL), which addresses a different problem. Rather than merely estimating the base expected value of a policy or for pre-specified set of candidate policies, OPL searches over a policy class $\Pi$ to find $\hat{\pi}^* = \arg\max_{\pi\in\Pi}\hat V(\pi)$, collapsing the evaluation into a single optimized output \cite{SNIPS,dudik_DR}. In this work, we deliberately do not perform OPL, mainly because  our scientific goal is \emph{comparative}: we wish to characterize the best available policy options among a diverse pre-specified portfolio of contextual bandit exploration strategies.

A parallel literature addresses uplift modeling, referring to the estimation of heterogeneous treatment effects (HTE), or conditional average treatment effects (CATE), expressed as $\tau(x)=\mathbb{E}[Y(1)-Y(0)\mid X=x]$,  where  $Y(1)$ and $ Y(0)$ are potential outcomes. Standard uplift evaluation relies on rank-based metrics such as the Qini coefficient and the area under the uplift curve (AUUC), which measure how well a model orders individuals by their individual treatment effect. While these metrics are natural for model selection, they do not directly capture the value of a deployed policy.

Here, we reframe the uplift problem from the model-centric AUUC perspective to a \emph{policy-value} perspective, measuring expected reward $V(\pi)$ and regret $\Delta(\pi)=V(\pi^\star)-V(\pi)$ relative to the best achieved policy $\pi^\star$. For doing that, three open datasets serve as our empirical benchmarks. The \emph{Hillstrom Mine That Data} dataset \cite{hillstrom2008-minethatdata} is a canonical three-arm email marketing randomized experiment whose treatment effect heterogeneity, while real, is modest, particularly for conversion and spend outcomes, as documented by \cite{molak2023-uplift-odyssey}. The \emph{Criteo Uplift} dataset \cite{diemert2018criteo} is a large-scale benchmark characterized by severe treatment imbalance and rare conversion outcomes. The \emph{LaLonde/NSW} dataset \cite{lalonde1986evaluating} and its re-analysis by \cite{lalondedehejia2002propensity} provide a benchmark for causal inference methods in which the experimental ground truth is known, here reinterpreted through the lens of policy regret rather than matching estimator bias.

Our work is mainly related to three research streams. First, there is a large literature on off-policy evaluation (OPE) for contextual bandits, including inverse-propensity, self-normalized, doubly robust, and bias-variance controlled estimators, as well as methods for valid inference under adaptive data collection \cite{dudik_DR,SNIPS,wang_switch,su2020dros,hadad2021ci}. Second, there is growing interest in reproducible OPE benchmarking and controlled experimentation environments, including open-source  software packages such as the Open Bandit Pipeline (OBP) and its associated public bandit datasets, which expose the practical challenges of evaluating policies from logged bandit feedback and standardize empirical comparisons \cite{saito_obp}. Third, our use of Hillstrom, Criteo, and LaLonde connects to the uplift and heterogeneous treatment effect literature, which studies how treatment effects vary across individuals and typically evaluates models through uplift-oriented metrics \cite{diemert2018criteo,gutierrez2017uplift,rzepakowski2012uplift}.

Beyond estimator design, a complementary line of work emphasizes the role of synthetic and semi-synthetic environments for controlled benchmarking of contextual bandits and off-policy methods. Such environments make it possible to vary reward structure, delays, concept drift, and business constraints while retaining access to a known reward-generating process. The simulator module of the OBP ecosystem provides configurable contextual bandit environments and policy simulators, thereby bridging offline evaluation and online experimentation in a general-purpose benchmarking framework \cite{saito_obp,obp_simulator}. Later extensions of OBP further incorporate industry-relevant challenges such as delayed feedback, concept drift, reward design, and operational constraints \cite{vandenakker2022obp}. Broader empirical benchmarking efforts also share this motivation of understanding algorithm behavior under controlled but practically meaningful conditions \cite{bietti_bakeoff_2021}.

What distinguishes the present paper from previous work is that we combine these strands under a deployment-oriented perspective. Rather than focusing only on estimator accuracy, synthetic benchmark design, or uplift ranking in isolation, we ask whether historical data from fixed randomized experiments can be used to decide if adaptive experimentation is worth deploying at all, and which adaptive policy classes are promising candidates before live interaction. In this sense, our contribution is not merely another OPE benchmark, another synthetic simulator, or another uplift application, but an end-to-end decision-support framework: fixed-test logs are used to estimate counterfactual policy values, identify promising adaptive candidates, and then stress-test those same offline-selected policies in a controlled warm-start simulator prior to roll-out.

\section{Methodology}

We generate i.i.d.\ samples $(X_i,A_i,Y_i)$ with observed context $X_i$, binary action $A\in\{0,1\}$ and binary outcome $Y\in\{0,1\}$ under a balanced logging policy $b(A=1\mid X)\equiv 0.5$. For each context $x \in X$, the data-generating process is defined by the arm-specific response probabilities $ \mu_1(x)=\Pr(Y=1\mid A=1,X=x)$, $\mu_0(x)=\Pr(Y=1\mid A=0,X=x), $ and the oracle policy is $ \pi^\star(x)=\mathbb{I}\{\mu_1(x)\ge \mu_0(x)\}. $ We consider four representative HTE regimes.

\paragraph{No context.} This is the limiting case in which no personalization is possible. The response probabilities are constant,
\begin{equation} \label{eq:no_contex}
    \mu_1=0.60 \text{ and } \mu_0=0.55,
\end{equation}
so the optimal rule is also constant.

\paragraph{Categorical heterogeneity.} For a discrete covariate $x\in\{1,2,3\}$, we specify the piecewise response probabilities $\Pr(Y=1\mid A=a,x=x_i) \equiv p^1 _a (x_i) $, with
\begin{equation}\label{eq:cat}
    \left[ p^1 _0 (1), p^1 _1 (1),p^1 _0 (2), p^1 _1 (2),p^1 _0 (3), p^1 _1 (3) \right] = \left[ 0.1, 0.8, 0.5, 0.5, 0.8, 0.1 \right]
\end{equation}

so that treatment is beneficial in some categories, neutral in others, and harmful in the remainder. This setting mimics segment-level heterogeneity such as customer type, region, or income band.

\paragraph{Linear heterogeneity.} For a continuous covariate $x\in[0,1]$ with linear effect, we define
\begin{equation} \label{eq:linear}
    \Pr(Y=1\mid A=a,x)=\alpha_a+\beta_a x, \text{ with } (\alpha_0,\beta_0,\alpha_1,\beta_1)=(0.9,-0.8,0.1,0.8).
\end{equation}
This yields a monotone switching structure in which the preferred treatment depends on the covariate value.

\paragraph{Oscillatory heterogeneity.} For a continuous covariate $x\in[0,1]$ with oscillatory effect, we define
\begin{equation}\label{eq:osc}
    \Pr(Y=1\mid A=a,x)= \frac12\left[1+\cos\!\big(2\pi(\alpha_a+\beta_a x)\big)\right],
\end{equation}
$$
\text{ with } (\alpha_0,\beta_0,\alpha_1,\beta_1)=(0,1,0.5,1).
$$
This yields alternating treatment advantage across the covariate range and captures periodic or seasonal heterogeneity.

\paragraph{Two covariates.} To study richer structure, we combine a linear and an oscillatory covariate. In the main experiments, we use the average combination
\begin{equation}\label{eq:double}
\Pr(Y=1\mid A=a,x=(x_{\text{lin}},x_{\text{osc}})) =\frac12\big(\mu^{\text{lin}}_a(x_{\text{lin}}) +\mu^{\text{osc}}_a(x_{\text{osc}})\big),
\end{equation}

and the product combination

\begin{equation}\label{eq:double_prod}
\Pr(Y=1\mid A=a,x=(x_{\text{lin}},x_{\text{osc}})) =\mu^{\text{lin}}_a(x_{\text{lin}})  \times \mu^{\text{osc}}_a(x_{\text{osc}}),
\end{equation}
where $\mu^{\text{lin}}_a(x_{\text{lin}}) $ and $\mu^{\text{osc}}_a(x_{\text{osc}})$ are given by  \eqref{eq:linear} and \eqref{eq:osc}. These two combinations preserves the probability range and produces a smooth two-dimensional decision surface.

\begin{figure}[tbp] \centering
\begin{subfigure}[b]{0.48\textwidth} \centering \includegraphics[width=\textwidth]{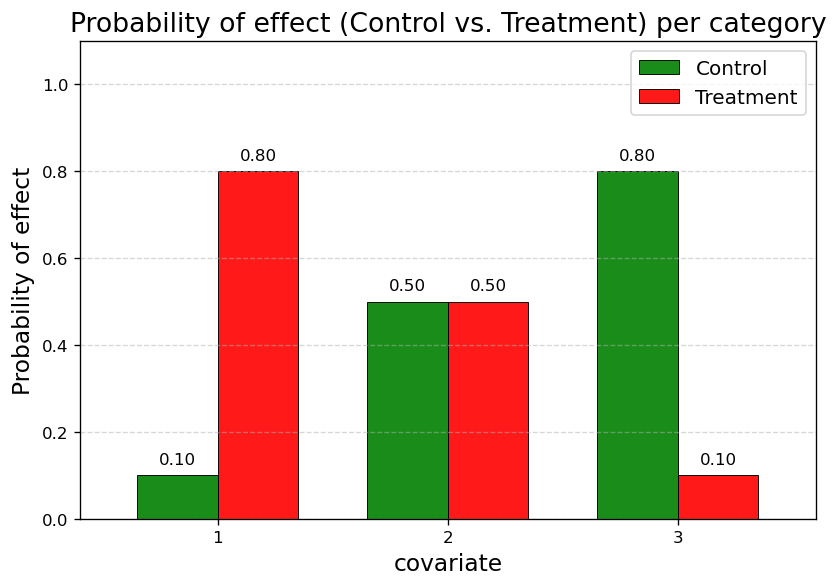} \caption{\textbf{Categorical heterogeneity.}} \label{fig:model_categorical} \end{subfigure}
\hfill \begin{subfigure}[b]{0.48\textwidth} \centering \includegraphics[width=\textwidth]{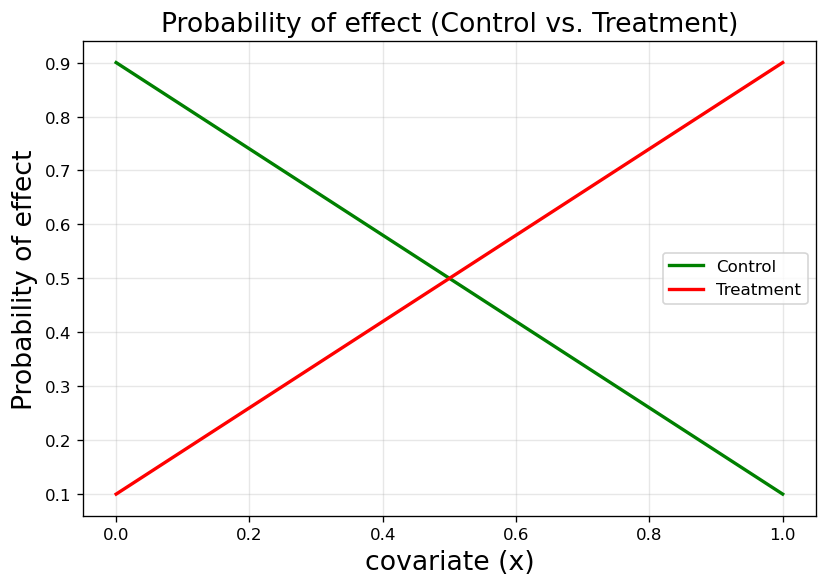} \caption{\textbf{Linear heterogeneity.}} \label{fig:model_linear} \end{subfigure}
\hfill \begin{subfigure}[b]{0.48\textwidth} \centering \includegraphics[width=\textwidth]{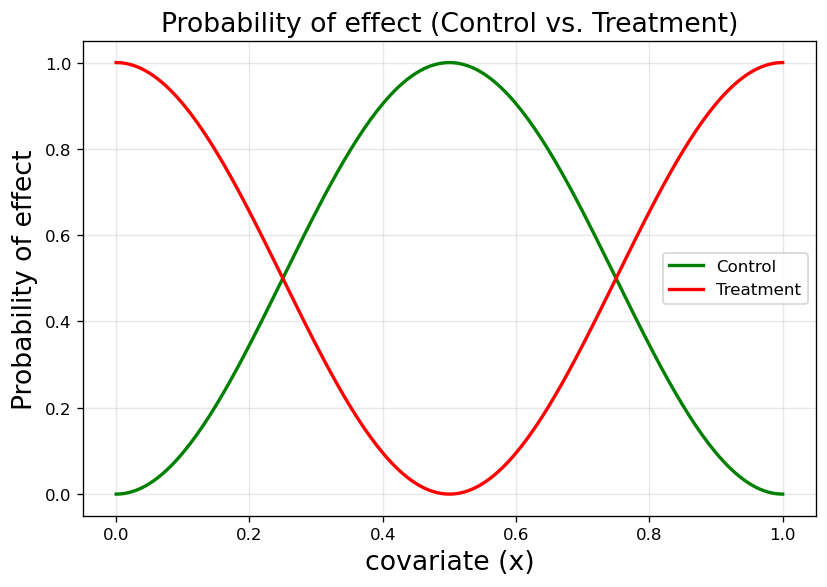} \caption{\textbf{Oscillatory heterogeneity.}} \label{fig:model_osc} \end{subfigure} 
\hfill \begin{subfigure}[b]{0.88\textwidth} \centering \includegraphics[width=\textwidth]{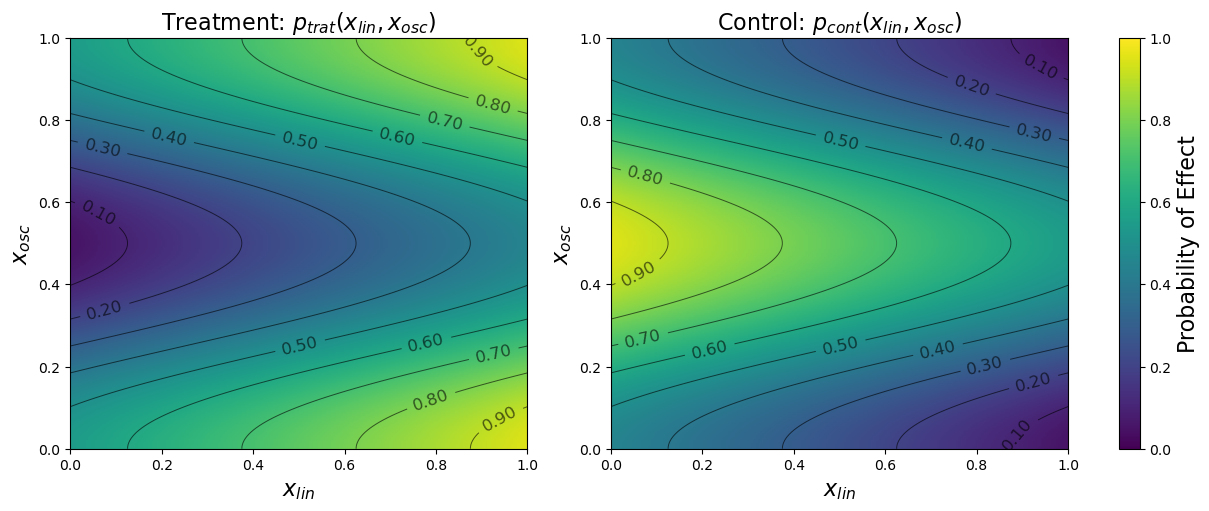} \caption{\textbf{Two covariates with average combination.}} \label{fig:model_2_covs_sum} \end{subfigure} 
\hfill \begin{subfigure}[b]{0.88\textwidth} \centering \includegraphics[width=\textwidth]{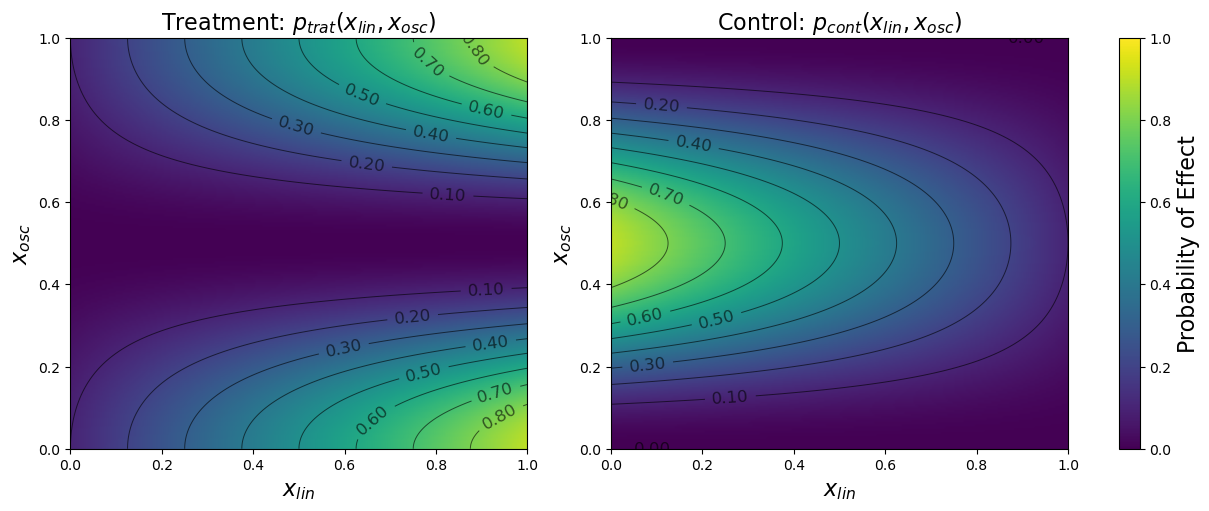} \caption{\textbf{Two covariates with product combination.}} \label{fig:model_2_covs_prod} \end{subfigure} 
\caption{Ground truth probability of effect given the Treatment and Covariate value. } \label{fig:models} \end{figure}

For OPE, we split the data so that the observations used to fit the nuisance components are disjoint from those used to evaluate candidate policies. This prevents optimistic bias and follows the standard sample-splitting logic used for estimation in \cite{dudik_DR,wang_switch}. We evaluate a fixed portfolio of contextual bandit policies (Bootstrapped Thompson Sampling, Bootstrapped UCB, Softmax/Boltzmann, $\varepsilon$-Greedy), together with fixed baselines and an oracle policy when $\mu_a(x)$ is known. On the held-out test split, we estimate policy value primarily with DR and use IPS and SNIPS as robustness checks \cite{dudik_DR,SNIPS,wang_switch}. Uncertainty is quantified via bootstrap confidence intervals for the OPE estimates. Inference for policy values in adaptive experiments is known to be delicate when assignment probabilities decay toward zero \cite{hadad2021ci}, but the synthetic OPE stage uses a fixed randomized logging policy with known propensities, so this extreme regime does not arise.

Each policy is initialized with the parameters learned during the offline stage and then deployed sequentially in this matched environment, receiving rewards sampled from the same $\mu_a(x)$ functions that generated the offline data. Accordingly, the simulator should be interpreted as a controlled best-case benchmark for offline-to-online transfer rather than as a realistic proxy for deployment under model misspecification, covariate shift, or temporal drift.

The OPE-predicted value $\hat V(\pi_k)$ serves as a pre-deployment forecast, but offline and online quantities are not identical objects in our setting. Offline OPE evaluates a stationary target rule $\pi(a\mid x)$, whereas the online bandit policies considered here (e.g., BootstrappedTS and BootstrappedUCB) are history-dependent and therefore induce learning-phase behavior that changes with past observations. Accordingly, discrepancies between offline OPE and realized online reward may arise not only from finite-sample noise and stochastic action selection, but also from this mismatch between stationary-policy evaluation and history-dependent online learning dynamics. For this reason, we interpret OPE primarily as a comparative diagnostic tool for screening policy classes and identifying promising candidates, rather than as a guarantee of exact online ranking among adaptive learners.

A key practical motivation for our pipeline is to provide a controlled environment in which offline-trained policies can be stress-tested before any live rollout. To this end, we construct a simulator that deliberately reuses the exact ground-truth reward functions $\mu_a(x)$. The online simulation proceeds as follows:

\begin{enumerate}%[label=\roman*)]
\item Draw $x_t$ i.i.d.\ from the same generator and compute an appropriate featurization $z_t=\phi(x_t)$;
\item For each OPE pre-trained policy $\pi$, choose $a_t\sim\pi(\cdot\mid z_t)$ (deterministic policies use $\arg\max$);
\item Sample $y_t\sim\!\left(\mu_{a_t}(x_t)\right)$;
\item Record instantaneous reward, cumulative reward, cumulative regret relative to the oracle, and action accuracy $\mathbb{I}[a_t=\pi^\star(x_t)]$;
\item Aggregate performance across replications and visualize policy-level trajectories and, when relevant, low-dimensional policy/value maps.
\end{enumerate}

These simulations constitute the final stage of the pipeline. By comparing realized cumulative reward and regret against both the oracle and the non-adaptive baselines, we measure how well candidate policies exploit the offline-learned structure when the online environment matches the offline data-generating process. Importantly, the online stage does not relearn a policy from logged feedback; it provides evidence about internal consistency and best-case offline-to-online transfer.

\section{Results} \label{sec:results}

We report results in two complementary layers. First, offline, we use Doubly Robust OPE to compare a portfolio of pre-specified adaptive and non-adaptive policies under a common policy-value and regret criterion. Second, when ground truth is available (synthetic data), we deploy the same offline-trained policies in the controlled warm-start simulator described above.

This two-stage design allows us to answer three practical questions:
\begin{enumerate}
\item whether adaptivity is beneficial relative to fixed allocation,
\item whether OPE-based coarse policy screening is preserved under finite-sample online deployment, and
\item whether the performance spread among strong adaptive candidates is large enough to justify additional offline policy optimization.
\end{enumerate}

Table \ref{tab:experiments} presents the Summary of Experiments that produces the results we report in this section. $N$: Number of observations in the fixed experiment (logging policy); $N_{Cov}$: Number of covariate columns; $N_{Boot}$: Number of Bootstraps done for OPE Confidence Intervals; $T_{On}$: Number of Online Simulation steps; $N_{On}$: Number of Online Simulations repetitions.

\begin{table}[tb]
\centering
\caption{Summary of Experiments ran for results generation.}
\label{tab:experiments}
\resizebox{\textwidth}{!}{%
\begin{tabular}{|l|l|l|l|l|l|l|l|l|}
\hline
Name & Ground Truth& $N$  &$N_{Cov}$ &Treatment/Control & $N_{Boot}$ & Online & $T_{On}$ & $N_{On}$ \\
 & & & & Allocation& & Simulation?& &\\
\hline
No Contex&     Eq. \eqref{eq:no_contex}          &     3000 &0 &50/50&            300&                    Yes&           500&           100\\
\hline
Categorical&   Eq. \eqref{eq:cat}           &     3000 &1 &50/50&            300&                    Yes&          500&          100\\
\hline
Linear &      Eq. \eqref{eq:linear}        &     3000 &1 &50/50&            300&                    Yes&          500&          100\\
\hline
Oscillatory &  Eq. \eqref{eq:osc}        &     3000 &1 &50/50&            300&                    Yes&          500&          100\\
\hline
Double Average&  Eq. \eqref{eq:double}        &     3000 &2 &50/50&            300&                    Yes&          500&          100\\
\hline
Double Product& Eq. \eqref{eq:double_prod}& 3000 &2 &50/50& 300& Yes& 500&100\\
\hline
Hillstrom &  None        &     64000&8 &33.3/33.3/33.3&            300&                    No&          - &          - \\
\hline
Criteo &  None        &     ~2.5m&11 &84.6/15.4&            300&                    No&          - &          - \\
\hline
Lalonde &  None        &     614&8 &30.1/69.9&            300&                    No&          - &          - \\
\hline
\end{tabular}%
}
\end{table}

\begin{table}[h]
\centering
\small
\caption{Structural differences between the two experiment families.}
\begin{tabular}{lll}
\toprule
\textbf{Aspect}            & \textbf{Synthetic}                        & \textbf{Open data}                      \\
\midrule
Data source                & Pre-generated RCT data& RCT or observational dataset                \\
Ground truth               & \texttt{FUNC\_TRAT}, \texttt{FUNC\_CONT}      & Not available                               \\
Oracle policy              & argmax(Ground Truth)& Not defined                                 \\
Logging propensity         & Uniform (RCT)                               & Uniform (RCT)                               \\
                           &                                               & or empirical frequency                      \\
Outcome model (binary)     & Calibrated RF per arm                         & Calibrated RF per arm                       \\
Outcome model (numeric)    & Ridge                                         & Ridge                                        \\
                           &/ HGB regressor per arm                        & / HGB regressor per arm                      \\
Covariate plots            & Yes: $\pi(x)$ and $\hat{V}_\pi(x)$ per bin   & No                                          \\
2-D heatmaps               & Yes (when 2-D covariates)                     & No                                          \\
Online simulation          & Yes ($N$ reps $\times$ $T$ steps)             & No                                          \\
\bottomrule
\end{tabular}
\end{table}

\subsection{No Context} \label{sec:res_no_context}

The no-context setting functions as a sanity check for the pipeline. Since no covariate information is available, the optimal rule is constant and there is no personalization problem to solve. Accordingly, the relevant question is not whether adaptive policies outperform through contextual adaptation, but whether they correctly recover the arm with the larger expected reward. Fig.\ref{fig:results_no_cov} shows that the strongest adaptive policies do identify the better arm and achieve strong OPE values, but this should not be interpreted as evidence that adaptivity is intrinsically useful in the absence of context. Rather, this case illustrates the limiting regime in which a strong fixed policy is already sufficient and any further policy optimization is unlikely to add practical value.

\subsection{Categorical Heterogeneity} \label{sec:res_categorical}

With discrete heterogeneity, adaptive policies consistently outperform non-adaptive baselines, as shown in Fig.\ref{fig:results_covariavel_categorica}. The important pattern is not that one specific adaptive policy dominates by a large margin, but that several adaptive candidates achieve similarly strong OPE values and low regret while the fixed baselines remain clearly inferior. This indicates that once category-level treatment differences are present, even relatively simple context-aware rules are sufficient to capture most of the available gain from personalization.

\subsection{Continuous and Structured Heterogeneity} \label{sec:res_continuous_structured}

\begin{figure}[tbp] \centering 
\begin{subfigure}[b]{0.48\textwidth} \centering \includegraphics[width=\textwidth]{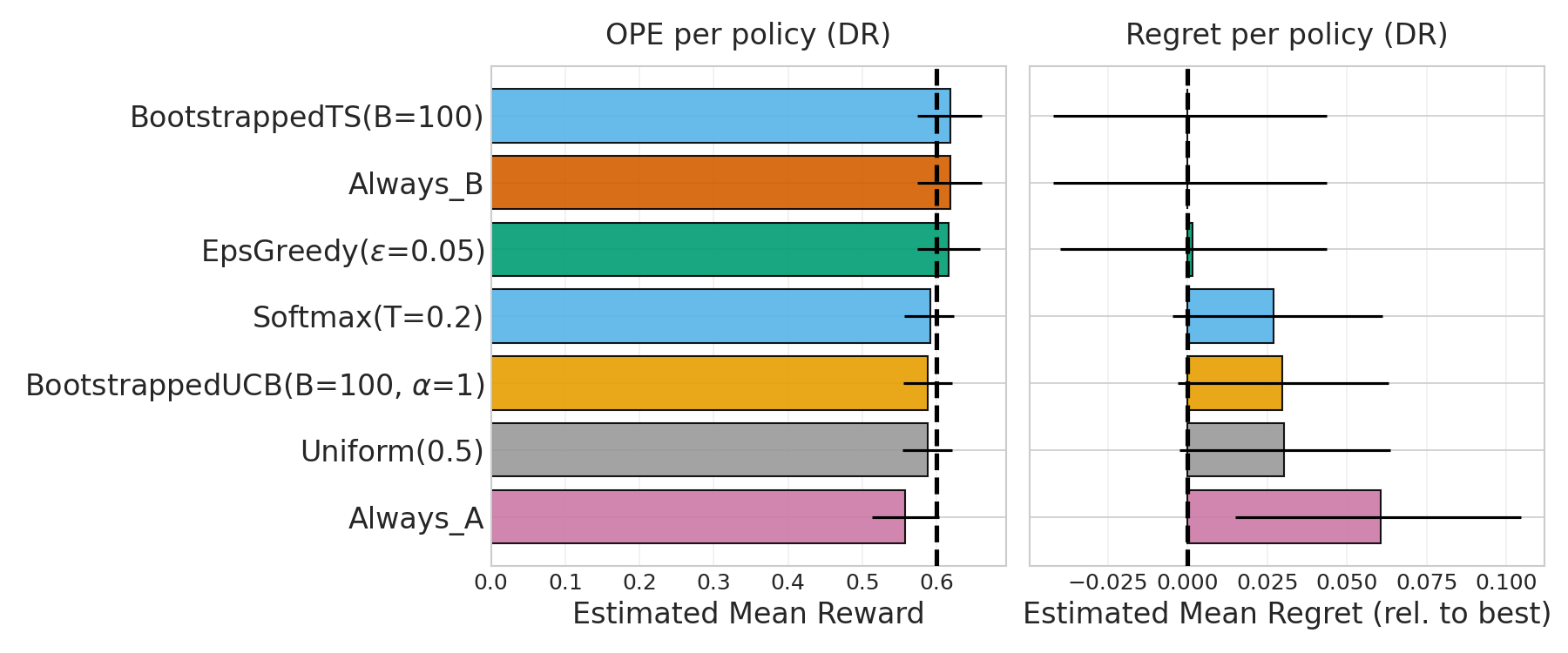} \caption{\textbf{No context.} } \label{fig:results_no_cov} \end{subfigure}
\hfill 
\begin{subfigure}[b]{0.48\textwidth} \centering \includegraphics[width=\textwidth]{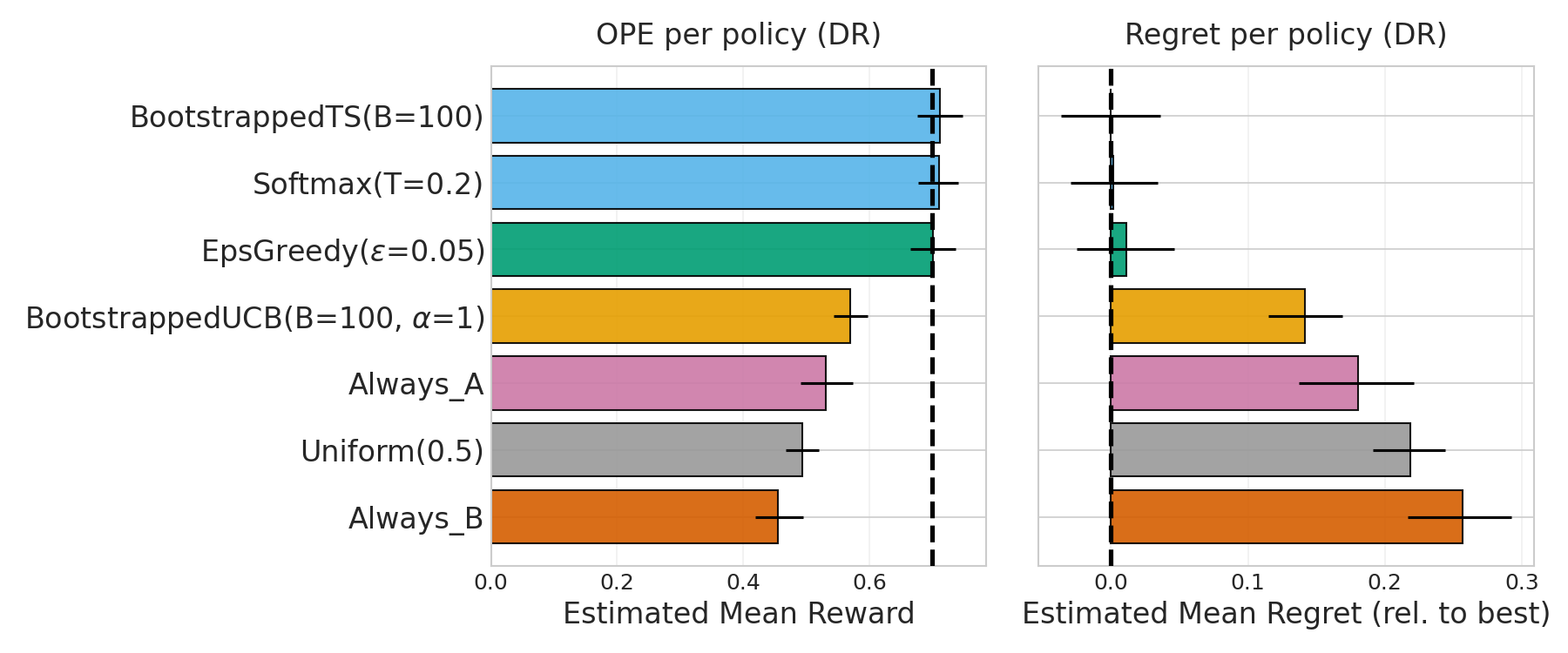} \caption{\textbf{Categorical heterogeneity.}} \label{fig:results_covariavel_categorica} \end{subfigure}
\hfill 
\begin{subfigure}[b]{0.48\textwidth} \centering \includegraphics[width=\textwidth]{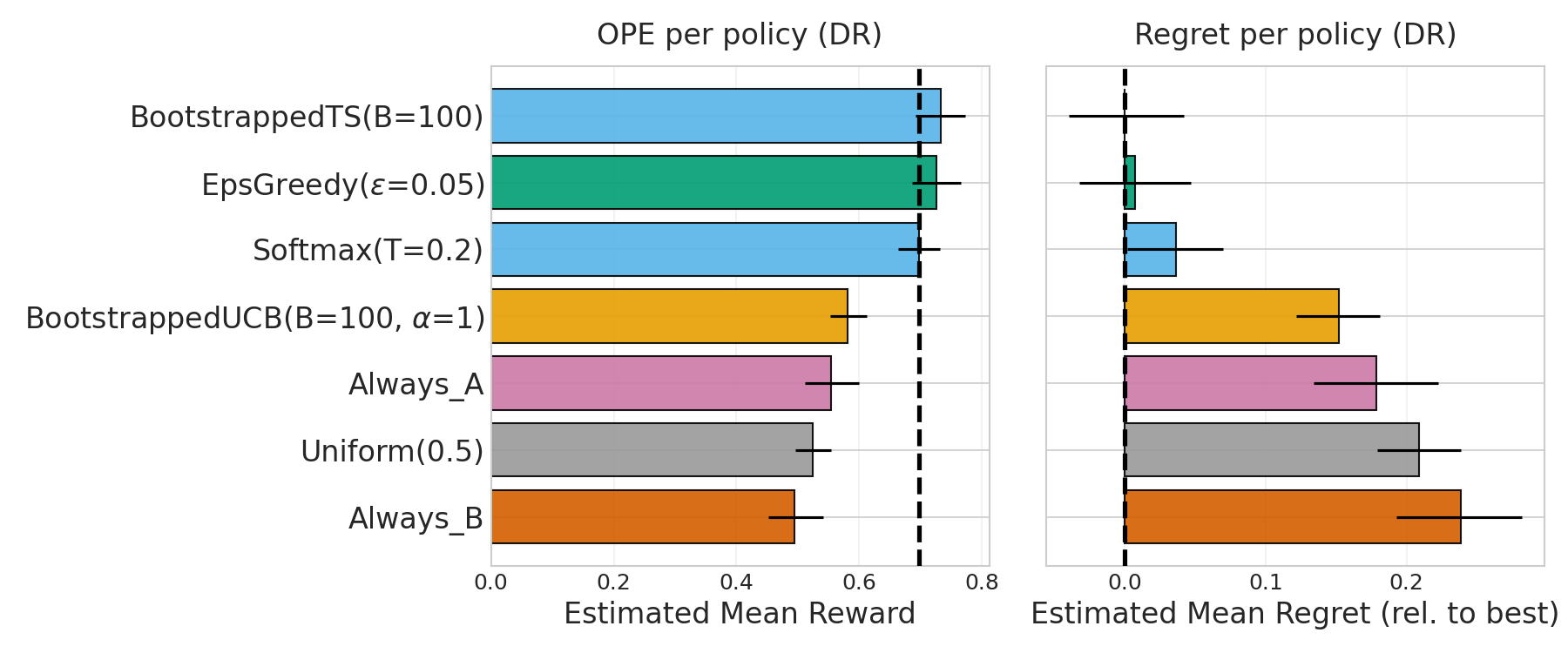} \caption{\textbf{Linear heterogeneity.}} \label{fig:results_covariavel_linear} \end{subfigure}
\hfill 
\begin{subfigure}[b]{0.48\textwidth} \centering \includegraphics[width=\textwidth]{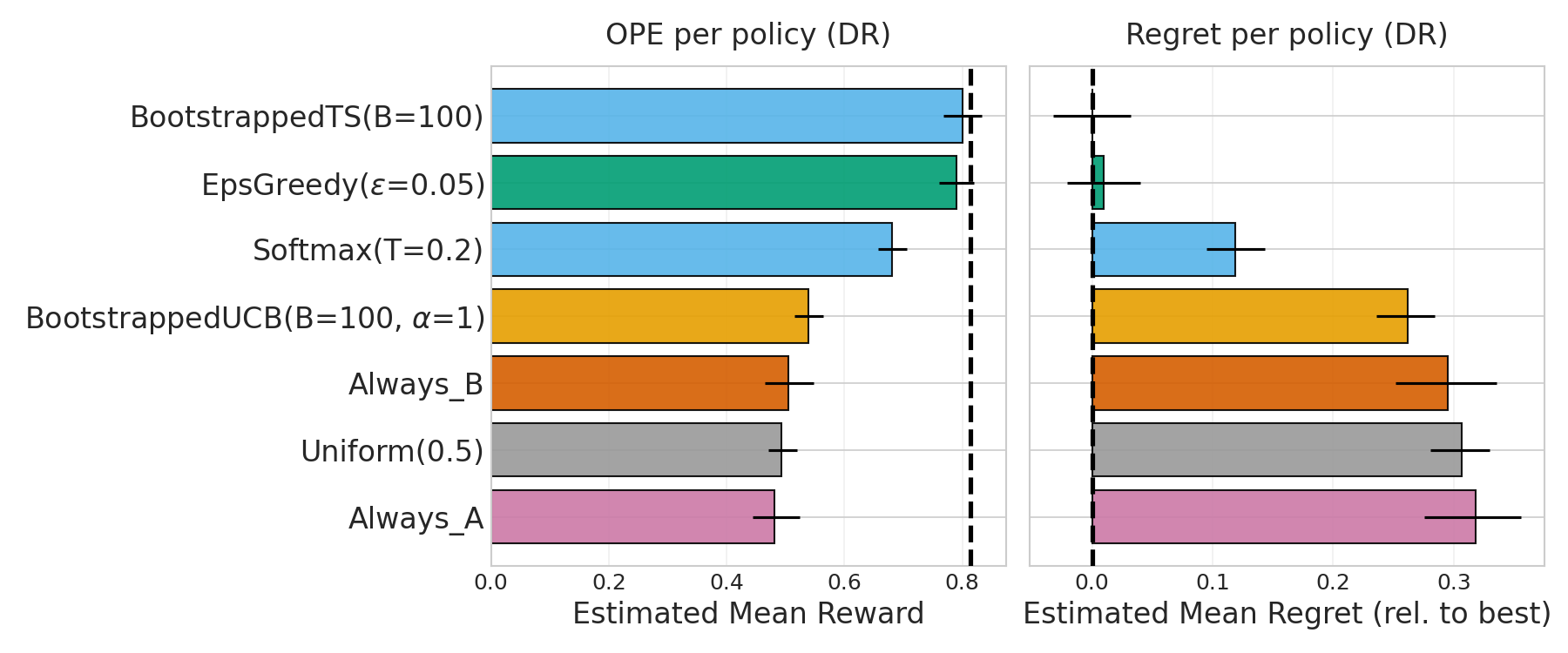} \caption{\textbf{Oscillatory heterogeneity.}} \label{fig:results_covariavel_oscilatoria} \end{subfigure} 
\hfill 
\begin{subfigure}[b]{0.48\textwidth} \centering \includegraphics[width=\textwidth]{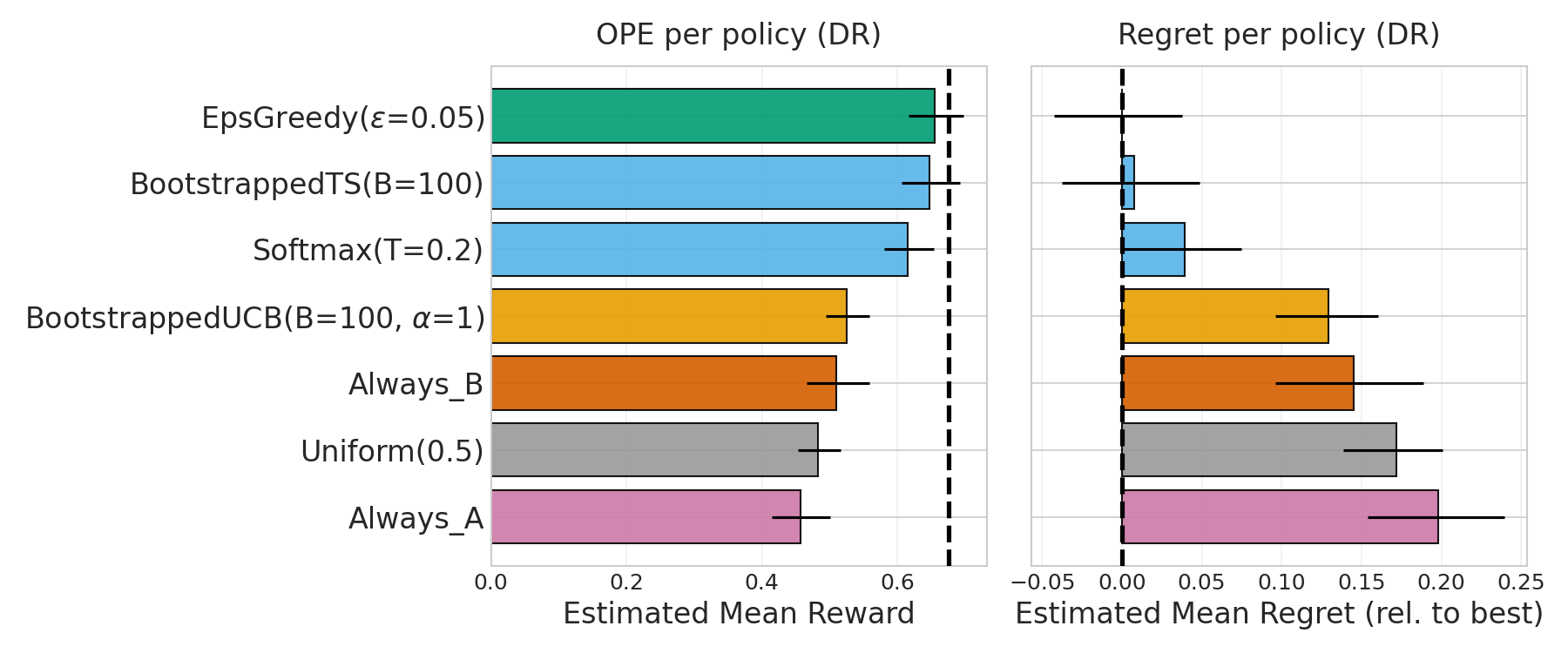} \caption{\textbf{Two covariates (average combination).}} \label{fig:results_covariavel_dupla_sum} \end{subfigure} 
\hfill 
\begin{subfigure}[b]{0.48\textwidth} \centering \includegraphics[width=\textwidth]{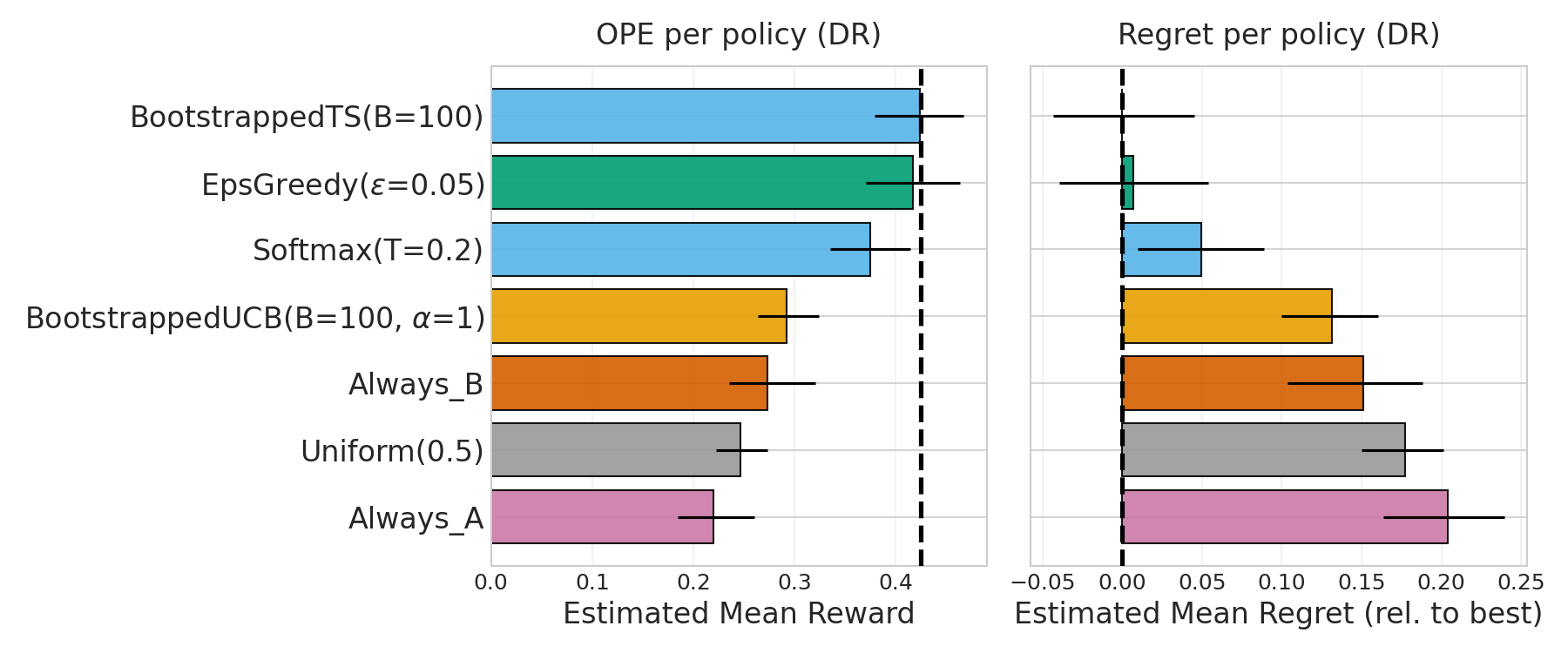} \caption{\textbf{Two covariates (product combination).}} \label{fig:results_covariavel_dupla_prod} \end{subfigure} \caption{\textbf{OPE (DR) and regret.} Policies are ranked by DR value (left) and estimated mean regret relative to the best observed policy (right). Whenever contextual heterogeneity is present, adaptive policies substantially outperform fixed baselines; among strong adaptive candidates, differences are comparatively modest.} \label{fig:results_ope_regret} \end{figure}

\begin{figure}[tbp] \centering
\begin{subfigure}[b]{0.60\textwidth} \centering \includegraphics[width=\textwidth]{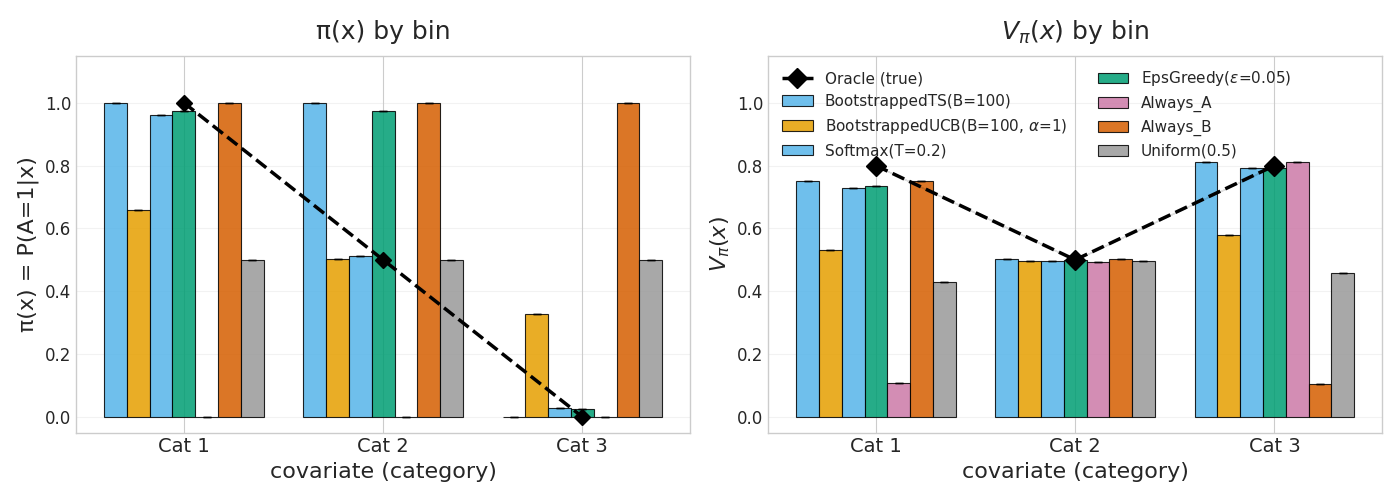} \caption{\textbf{Categorical heterogeneity.}} \label{fig:results2_covariavel_categorica} \end{subfigure}
\hfill \begin{subfigure}[b]{0.60\textwidth} \centering \includegraphics[width=\textwidth]{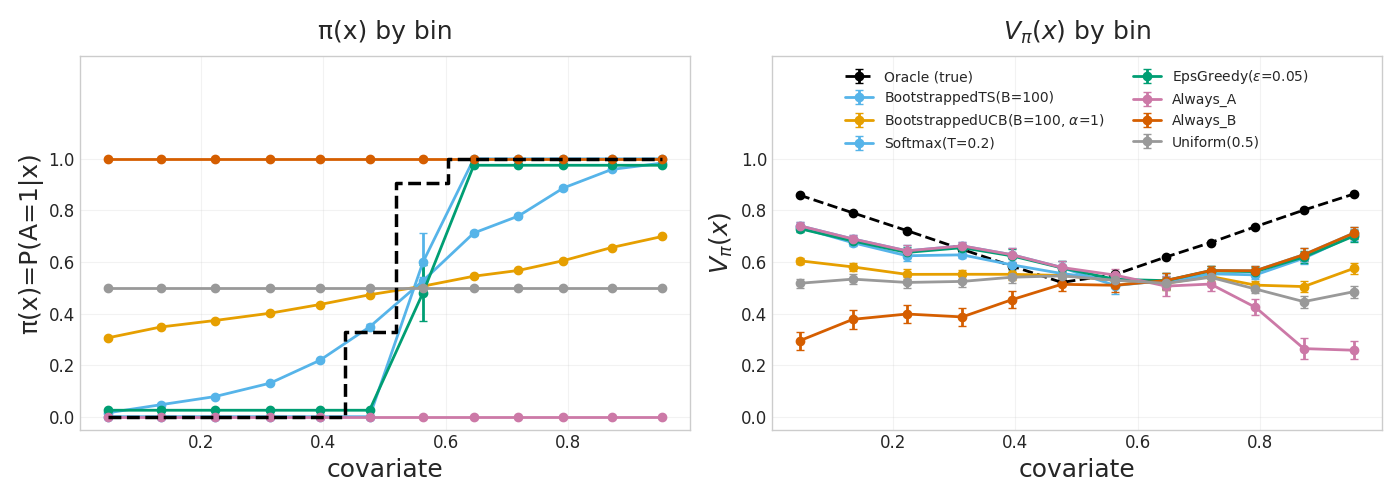} \caption{\textbf{Linear heterogeneity.}} \label{fig:results2_covariavel_linear} \end{subfigure}
\hfill \begin{subfigure}[b]{0.60\textwidth} \centering \includegraphics[width=\textwidth]{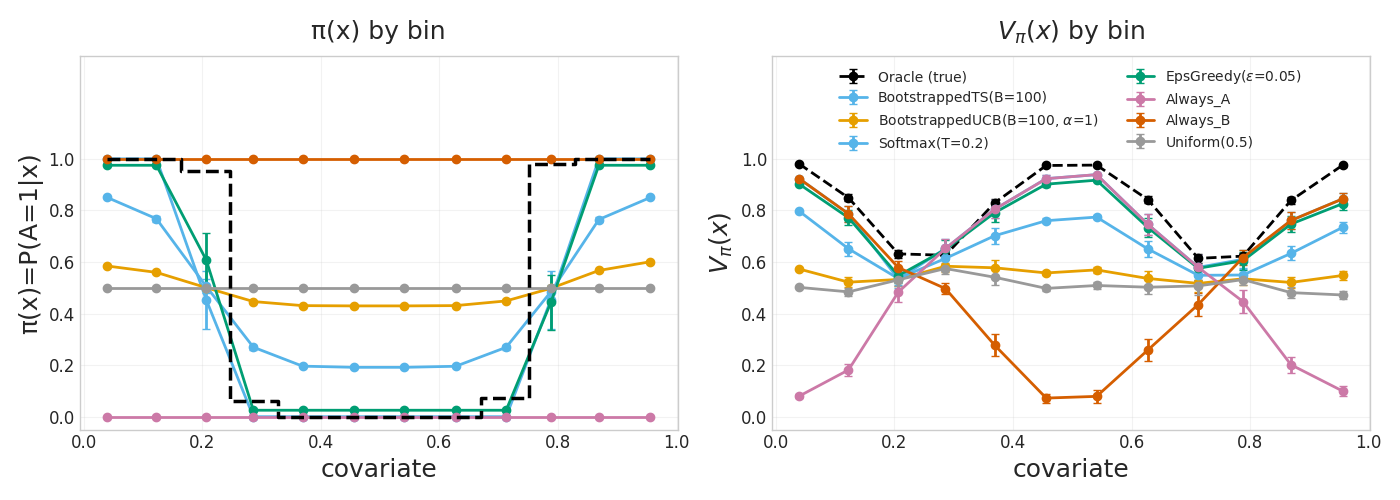} \caption{\textbf{Oscillatory heterogeneity.}} \label{fig:results2_covariavel_oscilatoria} \end{subfigure} \caption{\textbf{Policy and value functions by bins.} In the continuous settings, the learned policies recover the main decision structure implied by the oracle: a monotone switching threshold in the linear case and alternating treatment advantage in the oscillatory case. } \label{fig:results_vpi_pi} \end{figure}

Figs.\ref{fig:results_covariavel_linear}, \ref{fig:results_covariavel_oscilatoria}, and \ref{fig:results_covariavel_dupla_sum} summarize the offline OPE comparisons for the continuous and structured settings, while Figs.\ref{fig:results2_covariavel_linear} and \ref{fig:results2_covariavel_oscilatoria} visualize the learned policy and value structures. Online warm-start performance is reported in Table~\ref{tab:online}.

Across the linear (Figs.\ref{fig:results_covariavel_linear},\ref{fig:results2_covariavel_linear}), oscillatory (Figs.\ref{fig:results_covariavel_oscilatoria},\ref{fig:results2_covariavel_oscilatoria}), and two-covariate settings (Fig.\ref{fig:results_covariavel_dupla_sum}), the dominant result is stable: adaptive policies substantially outperform the fixed baselines, and the learned policy/value surfaces recover the structure implied by the oracle. In the linear case, the learned policies recover a monotone switching threshold; in the oscillatory case, they recover alternating treatment advantage across the covariate range; and in the two-covariate case, they assign probability mass to the regions favored by the oracle (full policy-surface plots are omitted for space). In all three settings, the strongest adaptive candidates cluster near the oracle benchmark, whereas the fixed baselines remain far behind. Thus, the practically important contrast is between adaptive and non-adaptive designs, not between the strongest adaptive candidates themselves.

\begin{figure}[tbp] \centering
\begin{subfigure}[b]{0.48\textwidth} \centering \includegraphics[width=\textwidth]{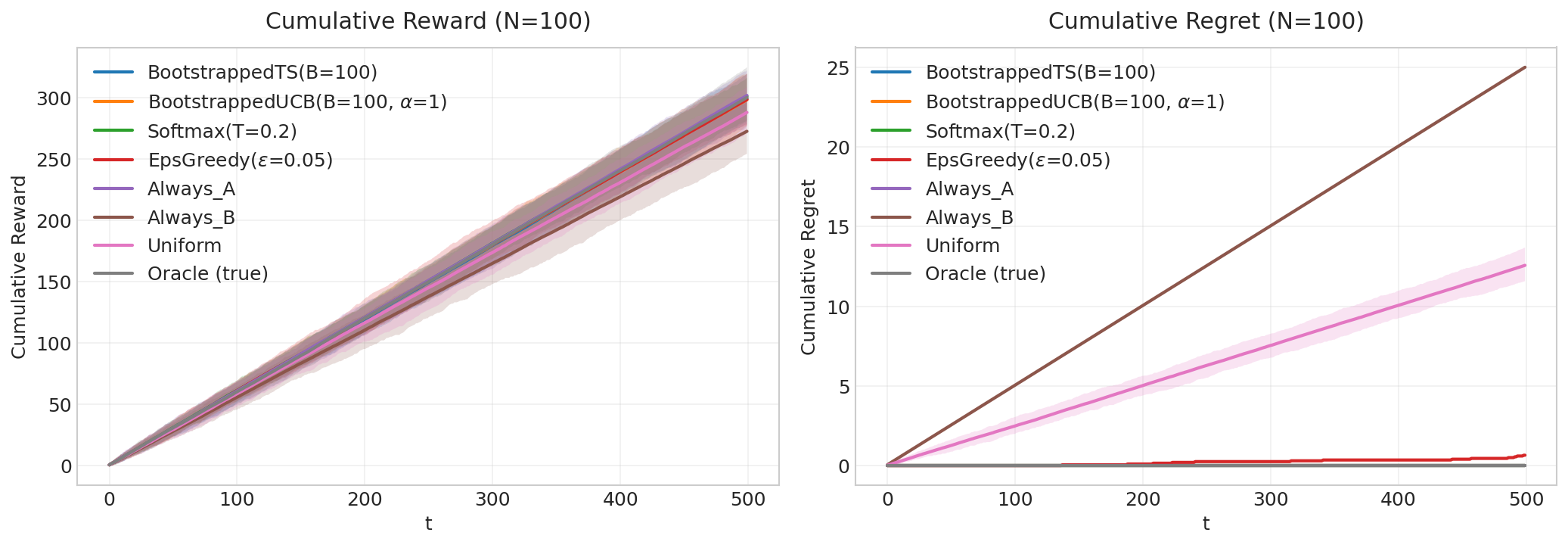} \caption{\textbf{No heterogeneity.} } \label{fig:results3_sem_covariavel} \end{subfigure}
\hfill \begin{subfigure}[b]{0.48\textwidth} \centering \includegraphics[width=\textwidth]{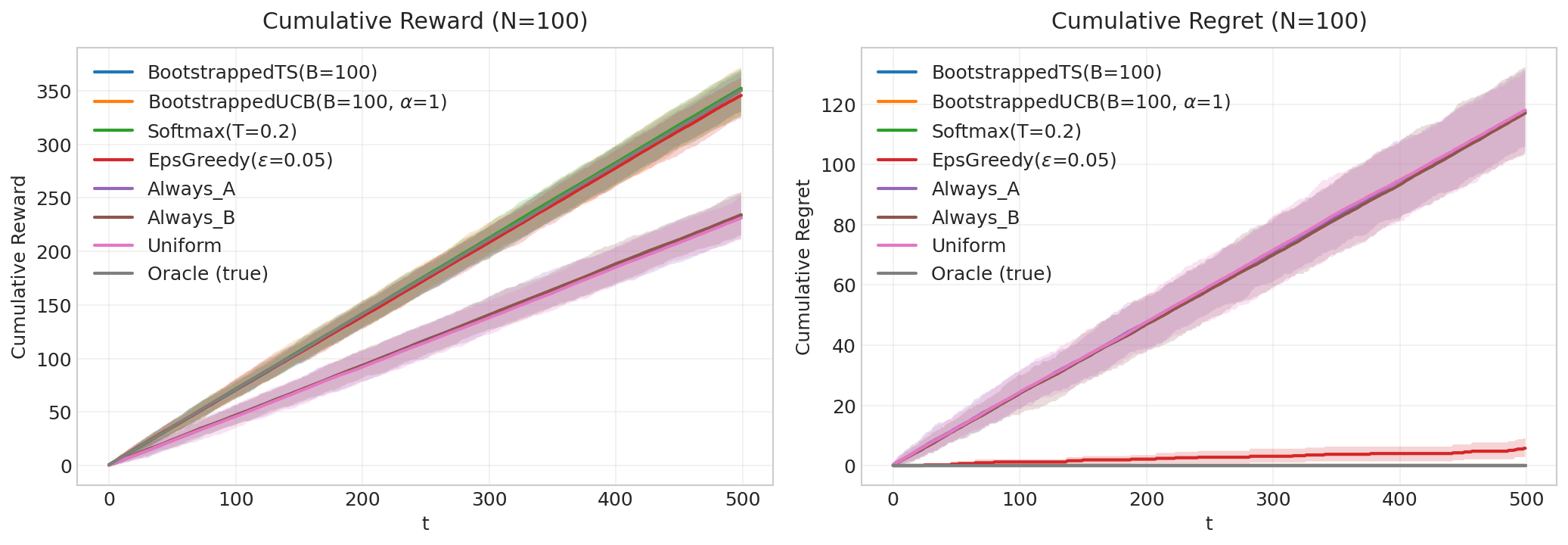} \caption{\textbf{Categorical heterogeneity.} } \label{fig:results3_covariavel_categorica} \end{subfigure}
\hfill \begin{subfigure}[b]{0.48\textwidth} \centering \includegraphics[width=\textwidth]{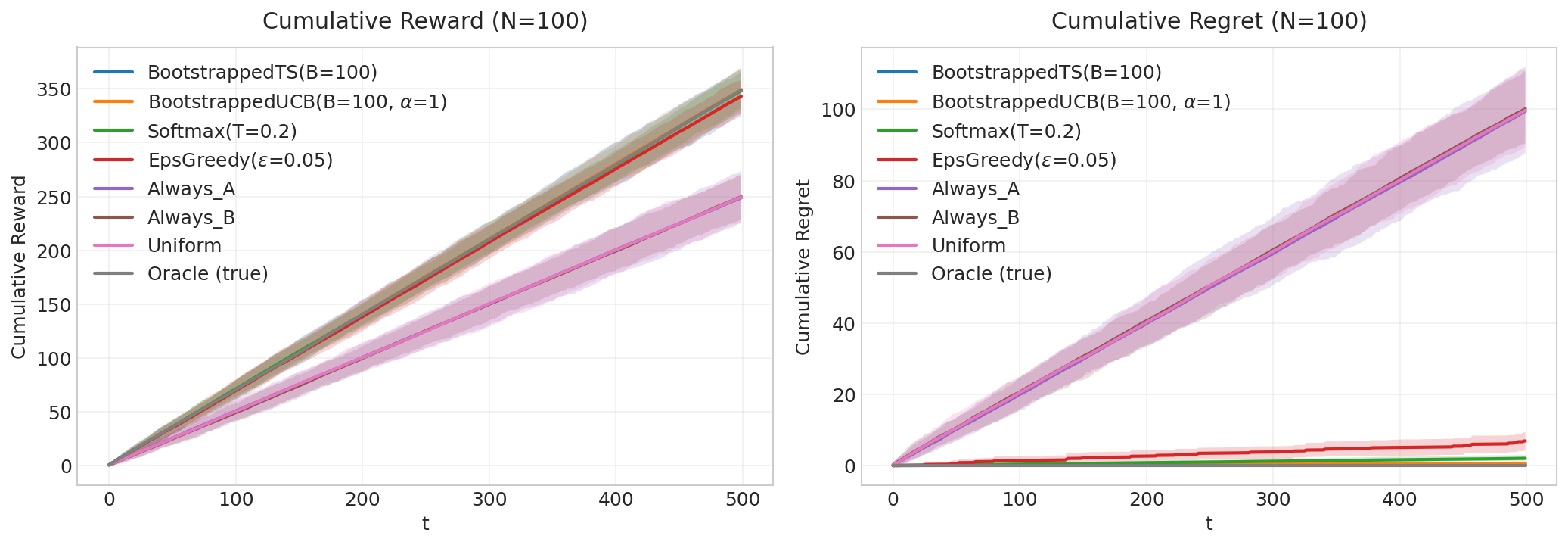} \caption{\textbf{Linear heterogeneity.} } \label{fig:results3_covariavel_linear} \end{subfigure}
\hfill \begin{subfigure}[b]{0.48\textwidth} \centering \includegraphics[width=\textwidth]{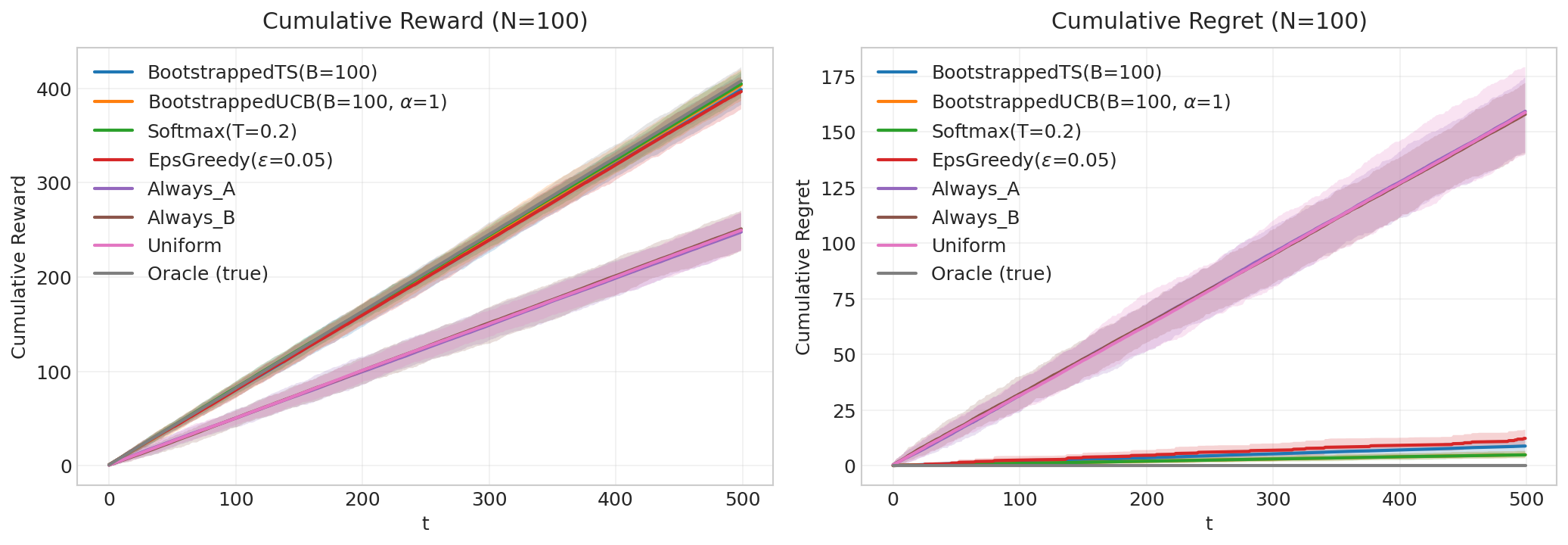} \caption{\textbf{Oscillatory heterogeneity.} } \label{fig:results3_covariavel_oscilatoria} \end{subfigure}
\hfill \begin{subfigure}[b]{0.48\textwidth} \centering \includegraphics[width=\textwidth]{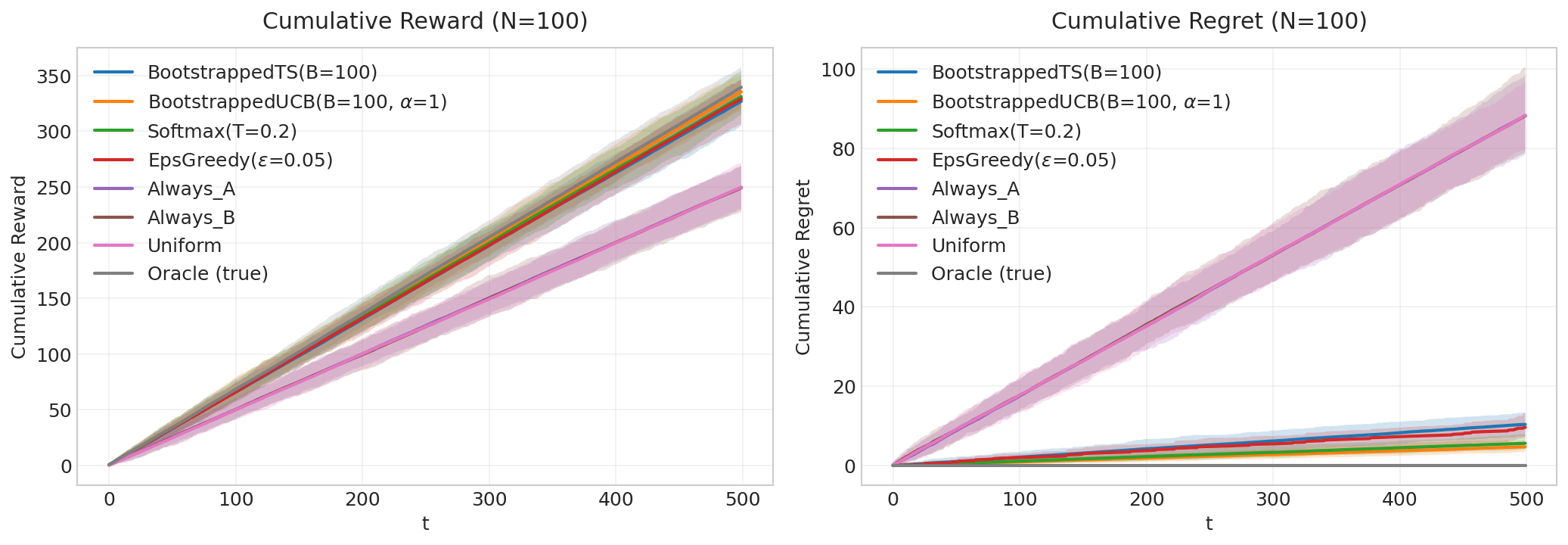} \caption{\textbf{Two covariates (average combination).} } \label{fig:results3_covariavel_dupla} \end{subfigure} 
\hfill \begin{subfigure}[b]{0.48\textwidth} \centering \includegraphics[width=\textwidth]{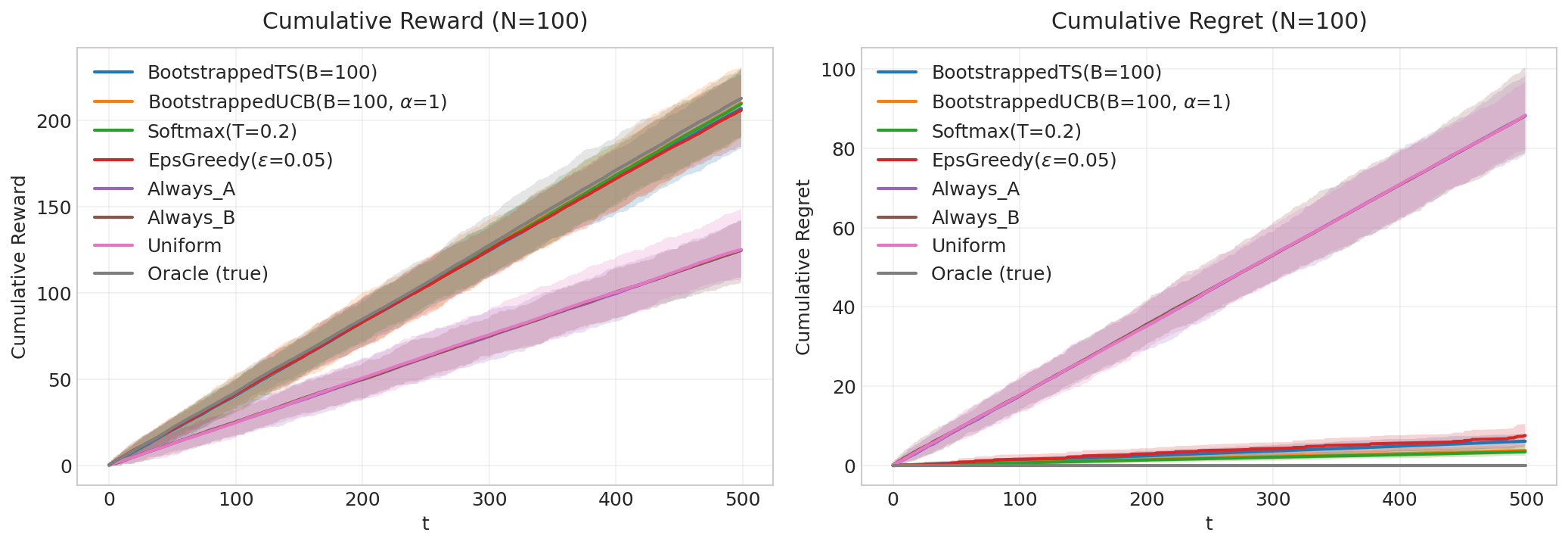} \caption{\textbf{Two covariates (product combination).} } \label{fig:results3_covariavel_dupla_prod} \end{subfigure} 
\caption{\textbf{Warm-start online cumulative metrics ($T{=}500, N{=}100$).} Left: cumulative reward; right: cumulative regret relative to the oracle. Several adaptive policies remain close to the oracle and far outperform the fixed baselines. The dominant contrast is between adaptive and non-adaptive designs; among strong adaptive policies, realized differences are modest and partly attributable to finite-sample variability.} \label{fig:results_online} \end{figure}

The online warm-start results reinforce this interpretation, as visible in Table ~\ref{tab:online} and in Figure \ref{fig:results_online}: the top adaptive policies remain close to the oracle across all structured HTE settings, while the fixed baselines accumulate substantially larger regret. When comparing the average and product combinations in the two-covariate setup, we do not observe meaningful differences in policy ordering or qualitative fit; the overall mean reward changes, but the main conclusions remain the same.

\begin{figure}[H] \centering 
\begin{subfigure}[b]{0.48\textwidth} \centering \includegraphics[width=\textwidth]{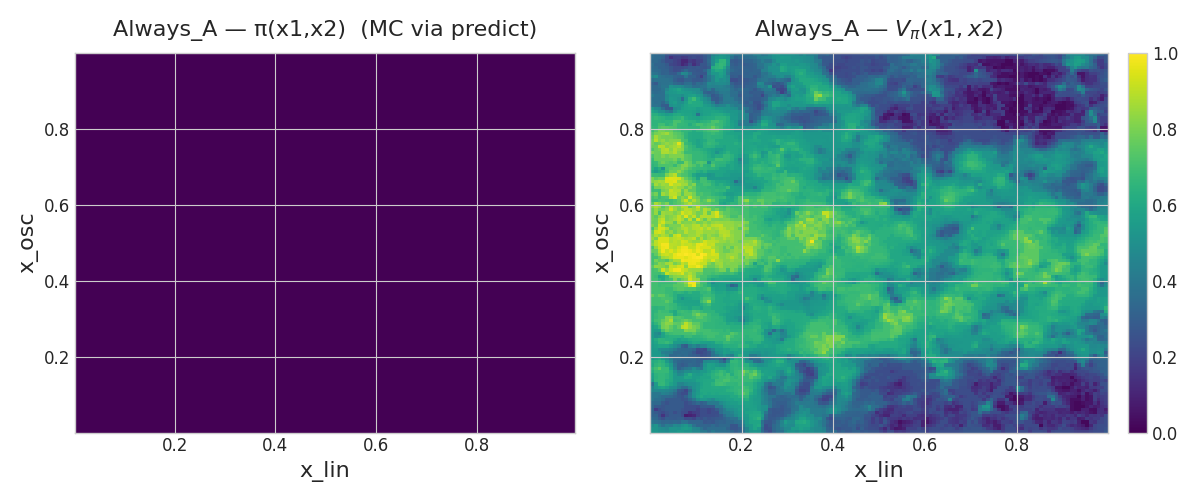} \caption{Always A policy.} \label{fig:2D_pi_vpi_Always_A_covariavel_dupla_sum} \end{subfigure} 
\hfill \begin{subfigure}[b]{0.48\textwidth} \centering \includegraphics[width=\textwidth]{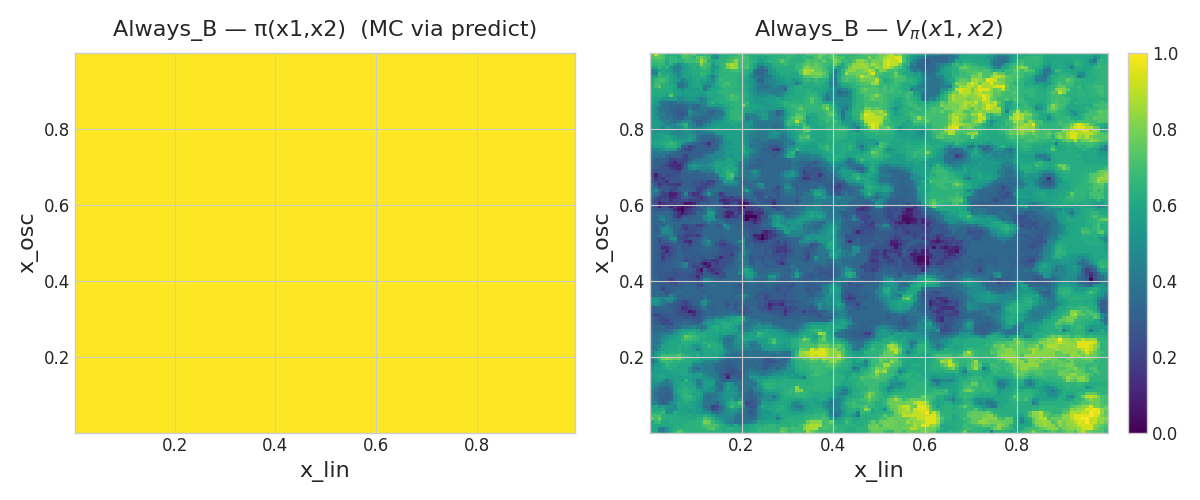} \caption{Always B policy.} \label{fig:2D_pi_vpi_Always_B_covariavel_dupla_sum} \end{subfigure} 
\hfill \begin{subfigure}[b]{0.48\textwidth} \centering \includegraphics[width=\textwidth]{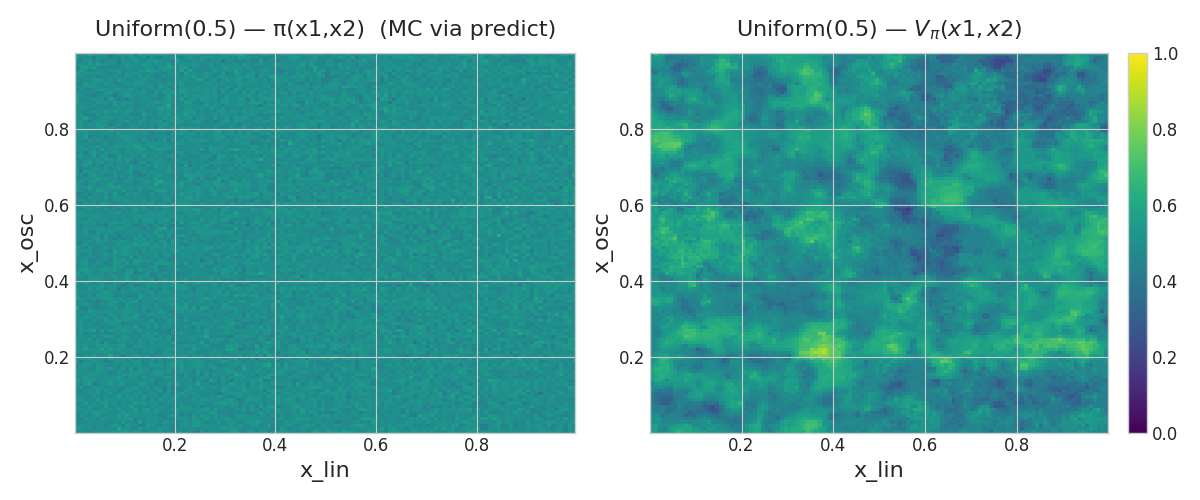} \caption{Uniform policy.} \label{fig:2D_pi_vpi_Uniform_covariavel_dupla_sum} \end{subfigure} 
\begin{subfigure}[b]{0.48\textwidth} \setcounter{subfigure}{3} \centering \includegraphics[width=\textwidth]{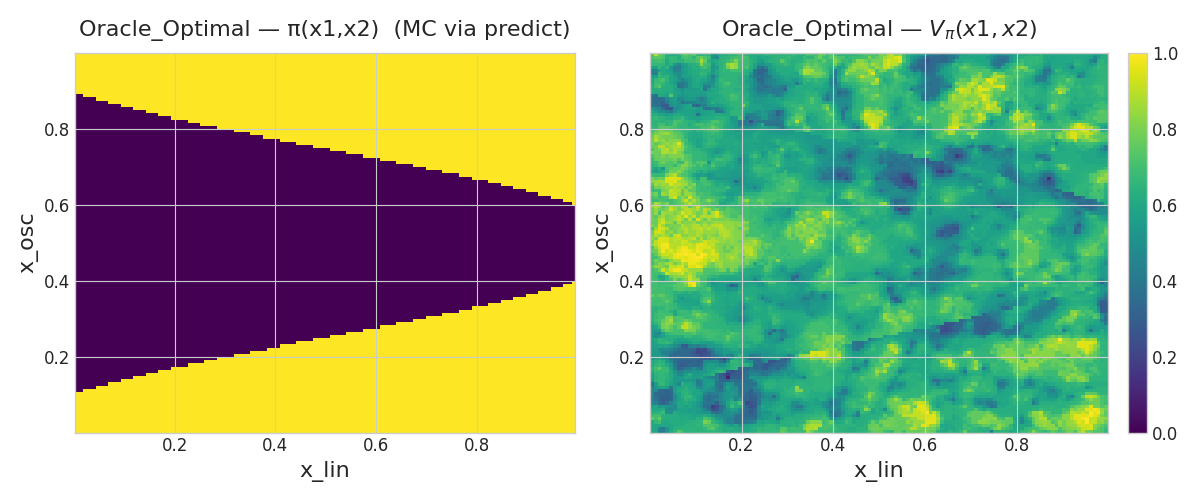} \caption{Oracle policy.} \label{fig:2D_pi_vpi_Oracle_covariavel_dupla_sum} \end{subfigure} 
\hfill \begin{subfigure}[b]{0.48\textwidth} \centering \includegraphics[width=\textwidth]{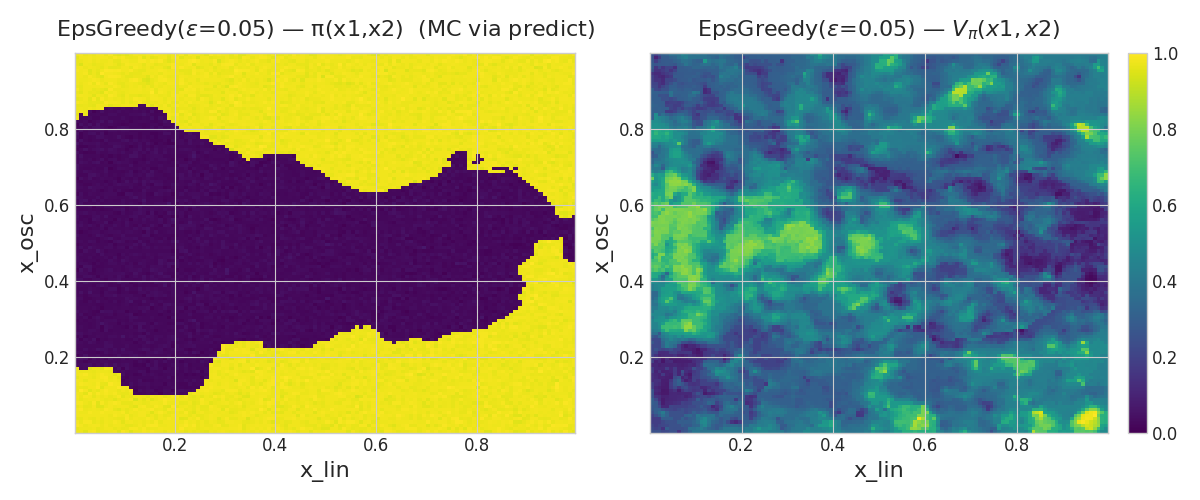} \caption{$\varepsilon$-Greedy policy.} \label{fig:2D_pi_vpi_EpsGreedy_covariavel_dupla_sum} \end{subfigure} 
%\hfill \begin{subfigure}[b]{0.48\textwidth} \centering \includegraphics[width=\textwidth]{img/2D_pi_vpi_Softmax_T=0.2__covariavel_dupla_sum_N=3000_a_tl=0.1_a_cl=0.9_b_tl=0.8_b_cl=-0.8_a_to=0_a_co=0.5_b_to=1_b_co=1.png} \caption{Softmax policy.} \label{fig:2D_pi_vpi_Softmax_covariavel_dupla_sum} \end{subfigure} 
%\begin{subfigure}[b]{0.48\textwidth} \setcounter{subfigure}{6} \centering \includegraphics[width=\textwidth]{img/2D_pi_vpi_BootstrappedUCB_covariavel_dupla_sum_N=3000_a_tl=0.1_a_cl=0.9_b_tl=0.8_b_cl=-0.8_a_to=0_a_co=0.5_b_to=1_b_co=1.png} \caption{Bootstrapped UCB policy.} \label{fig:2D_pi_vpi_BootstrappedUCB_covariavel_dupla_sum} \end{subfigure} 
\hfill \begin{subfigure}[b]{0.48\textwidth} \centering \includegraphics[width=\textwidth]{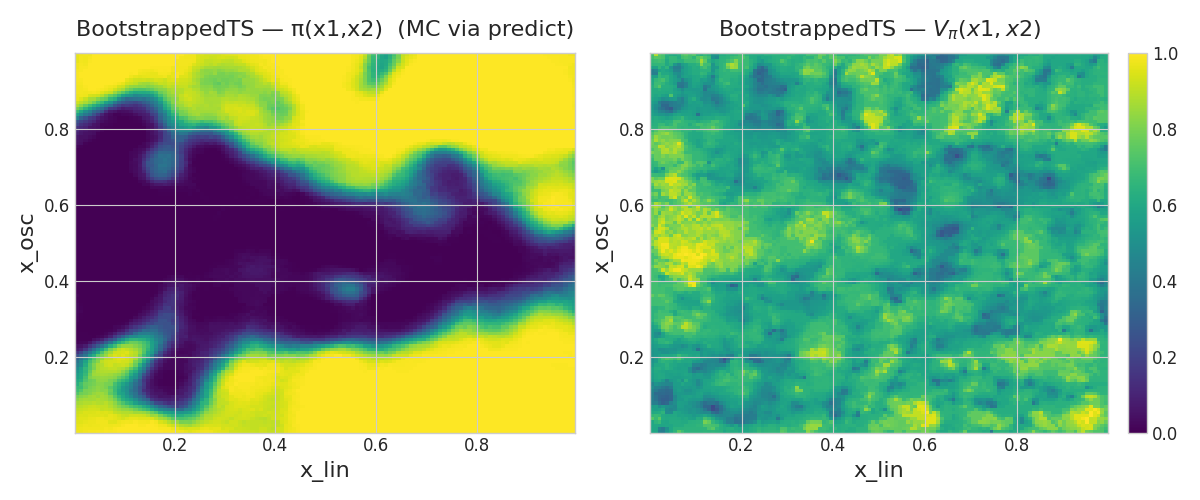}\caption{Bootstrapped Thompson Sampling policy.} \label{fig:2D_pi_vpi_BootstrappedTS_covariavel_dupla_sum} \end{subfigure} 
\caption{\textbf{Two-covariate with average combination policy and value surfaces.} Panels show $\pi(x_1,x_2)$ (left) and $V_\pi(x_1,x_2)$ (right) for each policy. Adaptive policies concentrate probability mass in regions favored by the oracle, while fixed baselines fail to exploit the two-dimensional heterogeneity structure. Softmax and UCB policies offers plots indistinguishable from the plots for Epsilon-Greedy policy.} \label{fig:2D_pi_vpi_sum} \end{figure} 

\begin{figure}[H] \centering 
\begin{subfigure}[b]{0.48\textwidth} \centering \includegraphics[width=\textwidth]{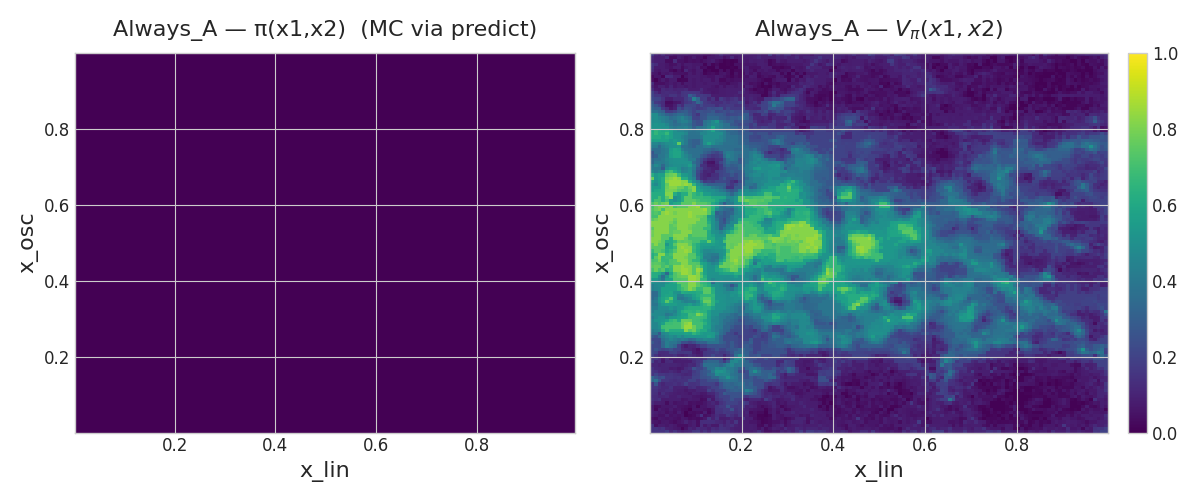} \caption{Always A policy.} \label{fig:2D_pi_vpi_Always_A_covariavel_dupla_prod} \end{subfigure} 
\hfill \begin{subfigure}[b]{0.48\textwidth} \centering \includegraphics[width=\textwidth]{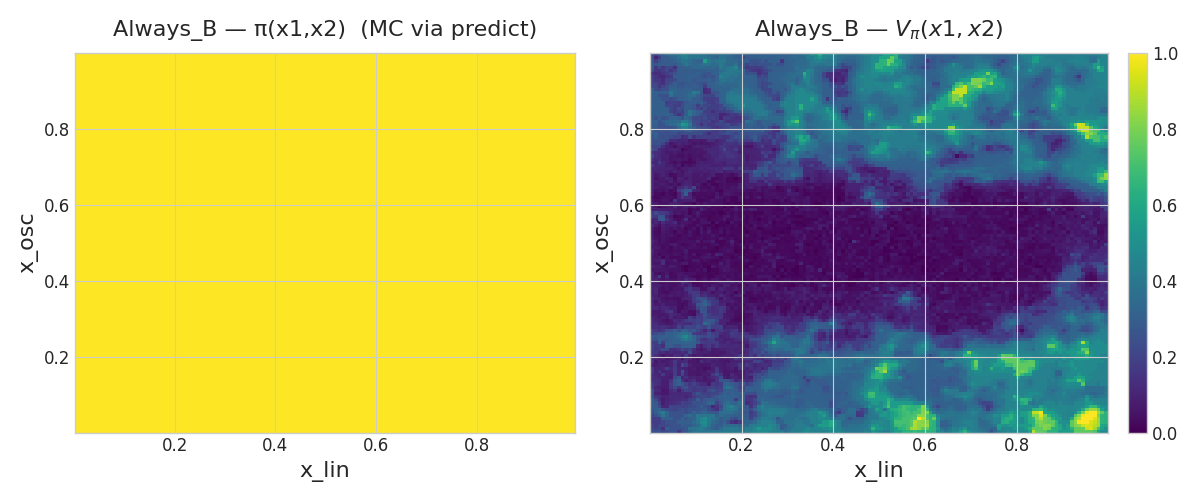} \caption{Always B policy.} \label{fig:2D_pi_vpi_Always_B_covariavel_dupla_prod} \end{subfigure} 
\hfill \begin{subfigure}[b]{0.48\textwidth} \centering \includegraphics[width=\textwidth]{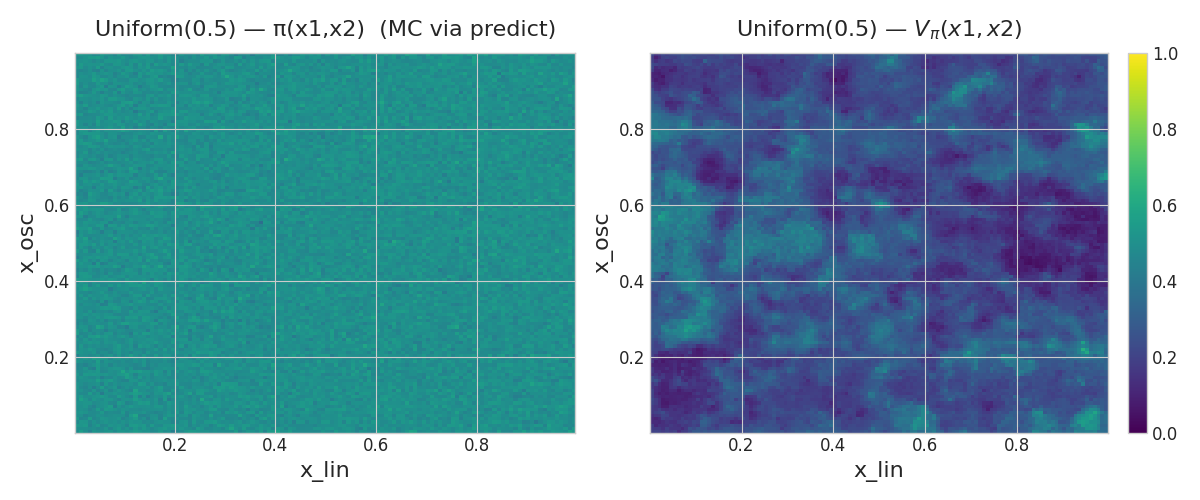} \caption{Uniform policy.} \label{fig:2D_pi_vpi_Uniform_covariavel_dupla_prod} \end{subfigure} 
\begin{subfigure}[b]{0.48\textwidth} \setcounter{subfigure}{3} \centering \includegraphics[width=\textwidth]{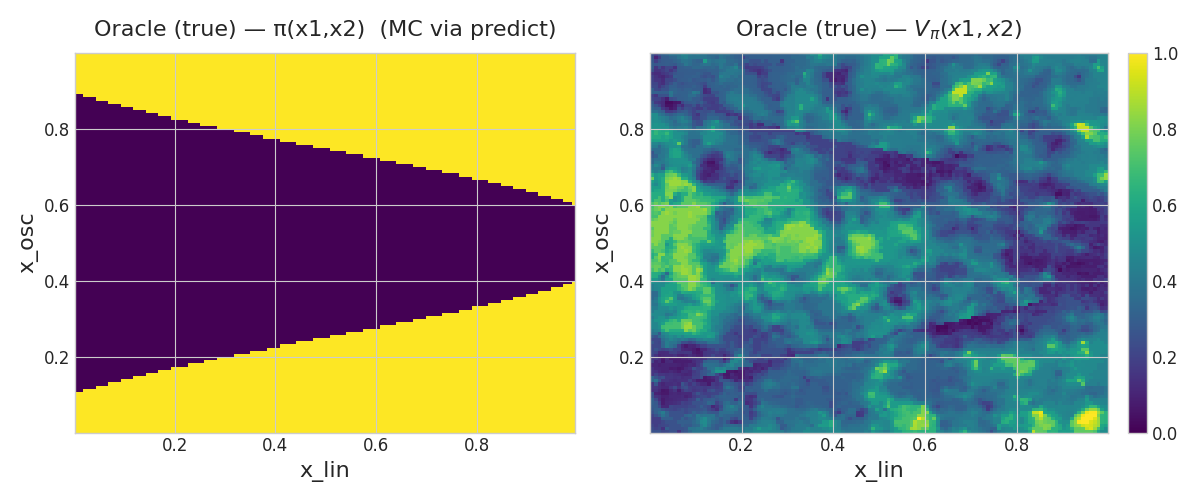} \caption{Oracle policy.} \label{fig:2D_pi_vpi_Oracle_covariavel_dupla_prod} \end{subfigure} 
\hfill \begin{subfigure}[b]{0.48\textwidth} \centering \includegraphics[width=\textwidth]{img/2D_pi_vpi_EpsGreedy__epsilon=0.05__covariavel_dupla_prod_N=3000_a_tl=0.1_a_cl=0.9_b_tl=0.8_b_cl=-0.8_a_to=0_a_co=0.5_b_to=1_b_co=1.png} \caption{$\varepsilon$-Greedy policy.} \label{fig:2D_pi_vpi_EpsGreedy_covariavel_dupla_prod} \end{subfigure} 
\hfill \begin{subfigure}[b]{0.48\textwidth} \centering \includegraphics[width=\textwidth]{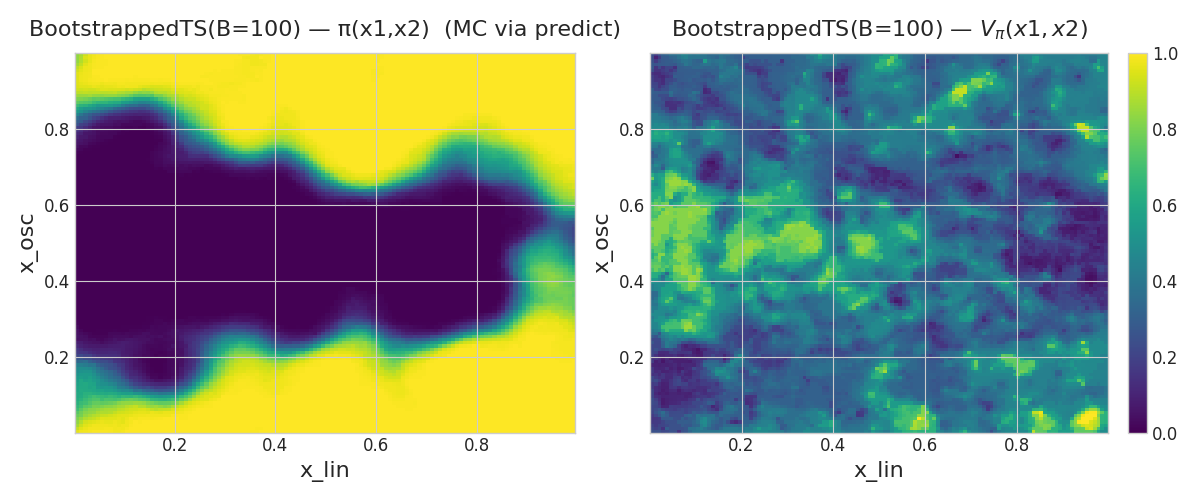}\caption{Bootstrapped Thompson Sampling policy.} \label{fig:2D_pi_vpi_BootstrappedTS_covariavel_dupla_prod} \end{subfigure} 
\caption{\textbf{Two-covariate with product combination policy and value surfaces.}} \label{fig:2D_pi_vpi_prod} \end{figure}

\begin{table}[tb]
\centering
\caption{\textbf{Warm-start online simulations ($T{=}500$ timesteps, average over $N=100$ simulations).} Accuracy is the fraction of oracle-consistent actions; reward and regret are reported both per-step and cumulatively. Across heterogeneous settings, strong adaptive policies remain close to the oracle and far outperform the fixed baselines.}
\label{tab:online}

\begin{subtable}[t]{0.32\textwidth} 
\centering 
\caption{No Covariates} 
\label{tab:online_nocov} 
\setlength{\tabcolsep}{2.5pt} 
\renewcommand{\arraystretch}{0.95} 
\scriptsize \resizebox{\linewidth}{!}{ 
\begin{tabular}{lcccc} 
\toprule 
Policy & Acc. & Mean R. & Cum. R. & Cum. Reg. \\ 
\midrule 
Always A & 1.000 & 0.603 & 301.55 & 0.00 \\ 
Softmax & 1.000 & 0.602 & 300.77 & 0.00 \\ 
Oracle & 1.000 & 0.600 & 299.89 & 0.00 \\ 
BootTS & 1.000 & 0.598 & 298.89 & 0.00 \\ 
$\varepsilon$-Greedy & 0.974 & 0.597 & 298.28 & 0.65 \\ 
BootUCB & 1.000 & 0.596 & 298.08 & 0.00 \\ 
Uniform & 0.497 & 0.575 & 287.71 & 12.57 \\ 
Always B & 0.000 & 0.545 & 272.33 & 25.00 \\ 
\bottomrule 
\end{tabular} } 
\end{subtable}
\hfill
\begin{subtable}[t]{0.32\textwidth} 
\centering 
\caption{Categorical} 
\label{tab:online_cat} 
\setlength{\tabcolsep}{2.5pt} 
\renewcommand{\arraystretch}{0.95} 
\scriptsize \resizebox{\linewidth}{!}{ 
\begin{tabular}{lcccc} 
\toprule 
Policy & Acc. & Mean R. & Cum. R. & Cum. Reg. \\ 
\midrule 
Softmax & 0.834 & 0.705 & 352.31 & 0.00 \\ 
BootUCB & 0.834 & 0.702 & 350.78 & 0.00 \\ 
BootTS & 0.837 & 0.701 & 350.58 & 0.00 \\ 
Oracle & 0.835 & 0.701 & 350.46 & 0.00 \\ 
$\varepsilon$-Greedy & 0.817 & 0.691 & 345.67 & 5.75 \\ 
Always B & 0.502 & 0.468 & 234.24 & 117.08 \\ 
Always A & 0.498 & 0.467 & 233.37 & 117.50 \\ 
Uniform & 0.498 & 0.463 & 231.27 & 118.06 \\ 
\bottomrule 
\end{tabular} } 
\end{subtable}
\hfill
\begin{subtable}[t]{0.32\textwidth}
\centering
\caption{Linear}
\label{tab:online_linear}
\setlength{\tabcolsep}{2.5pt}
\renewcommand{\arraystretch}{0.95}
\scriptsize
\resizebox{\linewidth}{!}{%
\begin{tabular}{lccccc}
\toprule
Policy & Acc. & Mean R. & Mean Reg. & Cum. R. & Cum. Reg. \\
\midrule
BootUCB    & 0.946 & 0.734 & 0.0016 & 367.0 & 0.80 \\
Oracle     & 1.000 & 0.730 & 0.0000 & 365.0 & 0.00 \\
Softmax    & 0.922 & 0.714 & 0.0042 & 357.0 & 2.08 \\
$\varepsilon$-Greedy & 0.912 & 0.694 & 0.0160 & 347.0 & 7.98 \\
BootTS     & 0.950 & 0.690 & 0.0027 & 345.0 & 1.34 \\
Always A   & 0.482 & 0.516 & 0.1941 & 258.0 & 97.07 \\
Always B   & 0.508 & 0.506 & 0.1848 & 253.0 & 92.40 \\
Uniform    & 0.492 & 0.506 & 0.1940 & 253.0 & 97.01 \\
\bottomrule
\end{tabular}%
}
\end{subtable}
\hfill
\begin{subtable}[t]{0.32\textwidth}
\centering
\caption{Oscillatory}
\label{tab:online_osc}
\setlength{\tabcolsep}{2.5pt}
\renewcommand{\arraystretch}{0.95}
\scriptsize
\resizebox{\linewidth}{!}{%
\begin{tabular}{lccccc}
\toprule
Policy & Acc. & Mean R. & Mean Reg. & Cum. R. & Cum. Reg. \\
\midrule
Oracle     & 1.000 & 0.852 & 0.0000 & 426.0 & 0.00 \\
BootUCB    & 0.934 & 0.830 & 0.0078 & 415.0 & 3.90 \\
$\varepsilon$-Greedy & 0.920 & 0.818 & 0.0158 & 409.0 & 7.88 \\
Softmax    & 0.908 & 0.804 & 0.0114 & 402.0 & 5.68 \\
BootTS     & 0.884 & 0.802 & 0.0332 & 401.0 & 16.59 \\
Uniform    & 0.530 & 0.510 & 0.2993 & 255.0 & 149.65 \\
Always B   & 0.516 & 0.502 & 0.3259 & 251.0 & 162.93 \\
Always A   & 0.436 & 0.470 & 0.3585 & 235.0 & 179.23 \\
\bottomrule
\end{tabular}%
}
\end{subtable}
\hfill
\begin{subtable}[t]{0.32\textwidth}
\centering
\caption{Two covariates with average combination}
\label{tab:online_two}
\setlength{\tabcolsep}{2.5pt}
\renewcommand{\arraystretch}{0.95}
\scriptsize
\resizebox{\linewidth}{!}{%
\begin{tabular}{lccccc}
\toprule
Policy & Acc. & Mean R. & Mean Reg. & Cum. R. & Cum. Reg. \\
\midrule
$\varepsilon$-Greedy & 0.860 & 0.720 & 0.0243 & 360.0 & 12.15 \\
Oracle     & 1.000 & 0.678 & 0.0000 & 339.0 & 0.00 \\
Softmax    & 0.924 & 0.672 & 0.0067 & 336.0 & 3.33 \\
BootUCB    & 0.924 & 0.652 & 0.0070 & 326.0 & 3.49 \\
BootTS     & 0.882 & 0.648 & 0.0163 & 324.0 & 8.13 \\
Uniform    & 0.496 & 0.492 & 0.1773 & 246.0 & 88.63 \\
Always A   & 0.454 & 0.482 & 0.1843 & 241.0 & 92.14 \\
Always B   & 0.516 & 0.480 & 0.1758 & 240.0 & 87.90 \\
\bottomrule
\end{tabular}%
}
\end{subtable}
\hfill
\begin{subtable}[t]{0.32\textwidth} 
\centering 
\caption{Two covariates with product combination}
\label{tab:online_two_prod} 
\setlength{\tabcolsep}{2.5pt}
\renewcommand{\arraystretch}{0.95}
\scriptsize
\resizebox{\linewidth}{!}{%
\begin{tabular}{lccccc}
\toprule 
Policy & Acc. & Mean R. & Cum. R. & Cum. Reg. \\ 
\midrule Oracle & 1.000 & 0.426 & 212.96 & 0.00 \\ 
BootUCB & 0.910 & 0.420 & 210.23 & 3.82 \\ 
Softmax & 0.912 & 0.420 & 209.79 & 3.48 \\ 
BootTS & 0.883 & 0.414 & 207.03 & 6.13 \\ 
$\varepsilon$-Greedy & 0.892 & 0.412 & 206.07 & 7.57 \\ 
Uniform & 0.501 & 0.251 & 125.28 & 88.27 \\ 
Always B & 0.498 & 0.250 & 124.98 & 88.13 \\ 
Always A & 0.502 & 0.250 & 124.81 & 88.22 \\ 
\bottomrule 
\end{tabular} } 
\end{subtable}
\end{table}

\subsection{Validating on Open Datasets}

To assess whether the same decision logic carries over beyond the synthetic experiments, we apply the OPE pipeline to three open benchmarks collected under fixed treatment assignment. Since no ground-truth reward surface is available in these datasets, we use only the offline stage of the framework. Before doing so, we generalize the implementation to handle binary and numeric outcomes, multiple treatment arms, and automatic covariate handling.

\begin{figure}[tbp]
\centering
\begin{subfigure}[b]{0.48\textwidth} \centering \includegraphics[width=\textwidth]{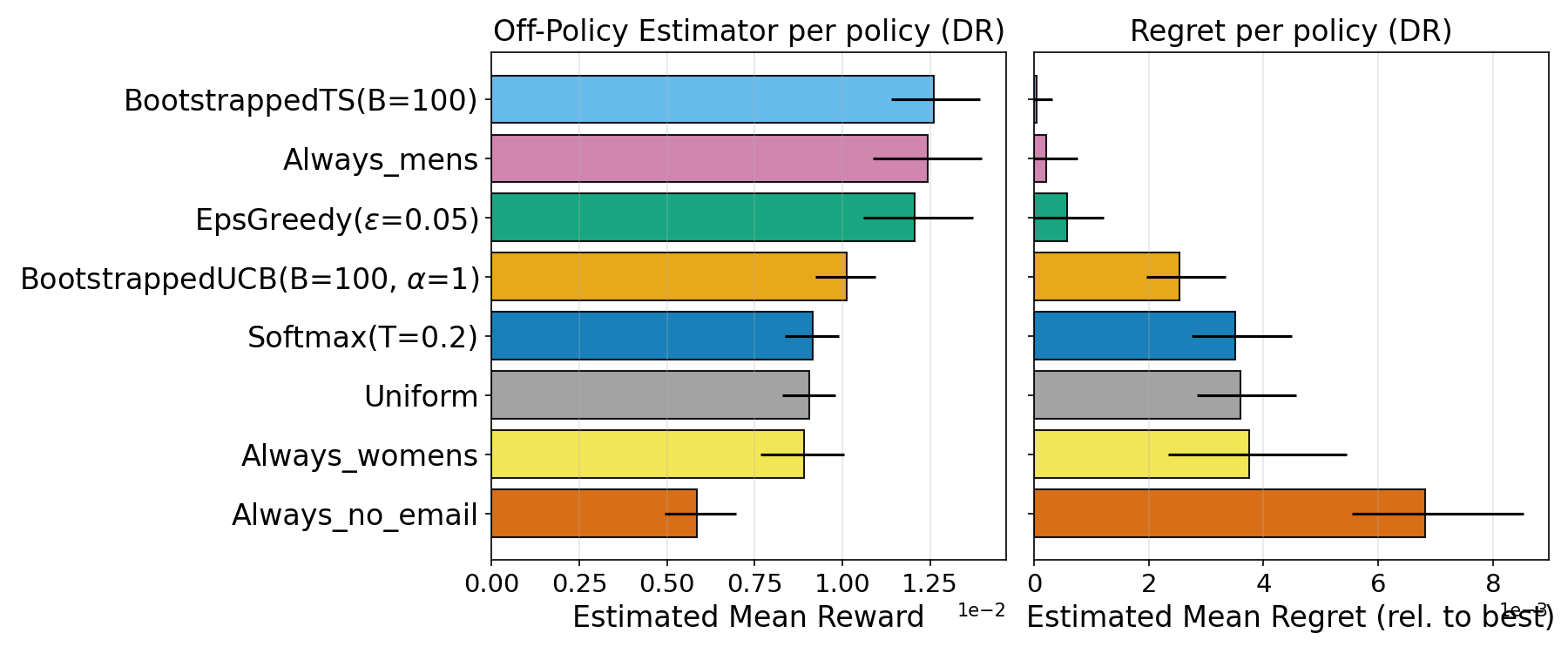} \caption{\textbf{Hillstrom Dataset} using conversion as outcome.} \label{fig:hillstorm_conversion} \end{subfigure}
\hfill\begin{subfigure}[b]{0.48\textwidth} \centering \includegraphics[width=\textwidth]{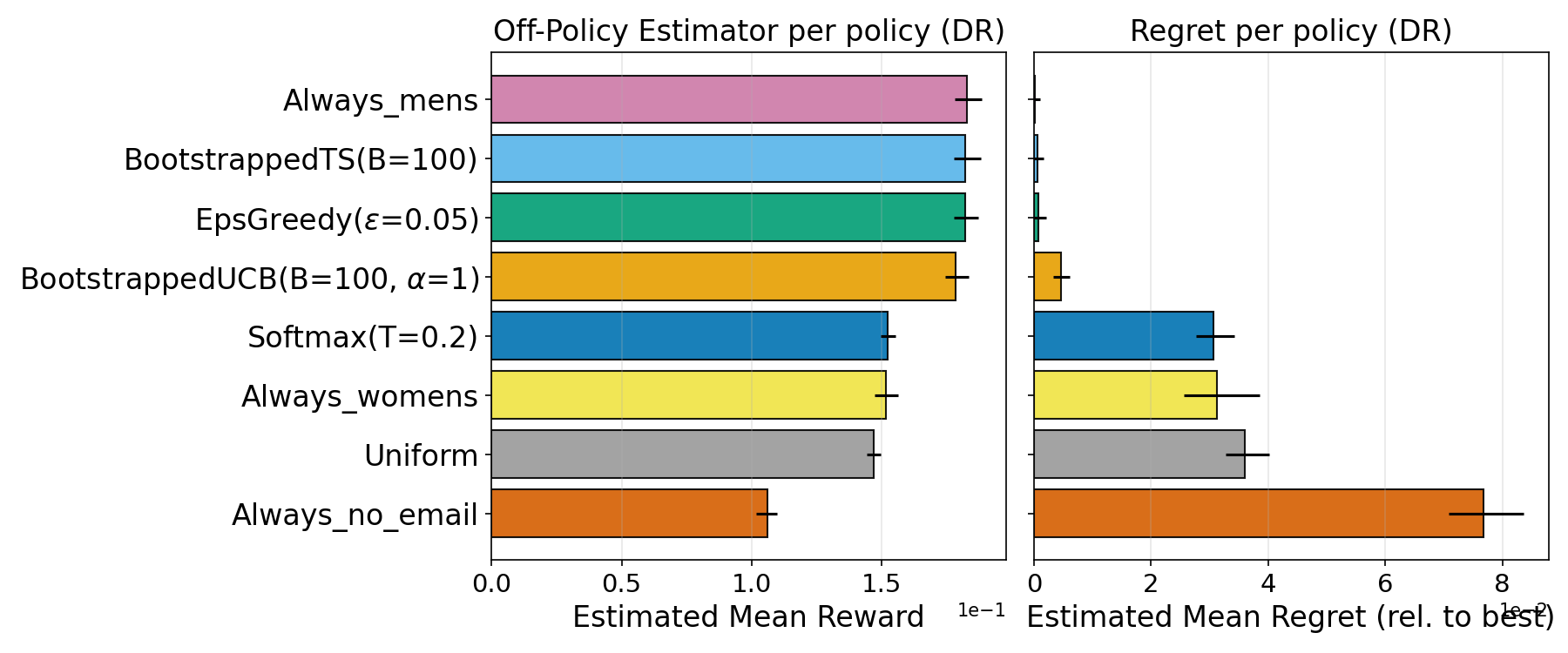} \caption{\textbf{Hillstrom Dataset} using visit as outcome.} \label{fig:hillstorm_visit} \end{subfigure}
\hfill\begin{subfigure}[b]{0.48\textwidth} \centering \includegraphics[width=\textwidth]{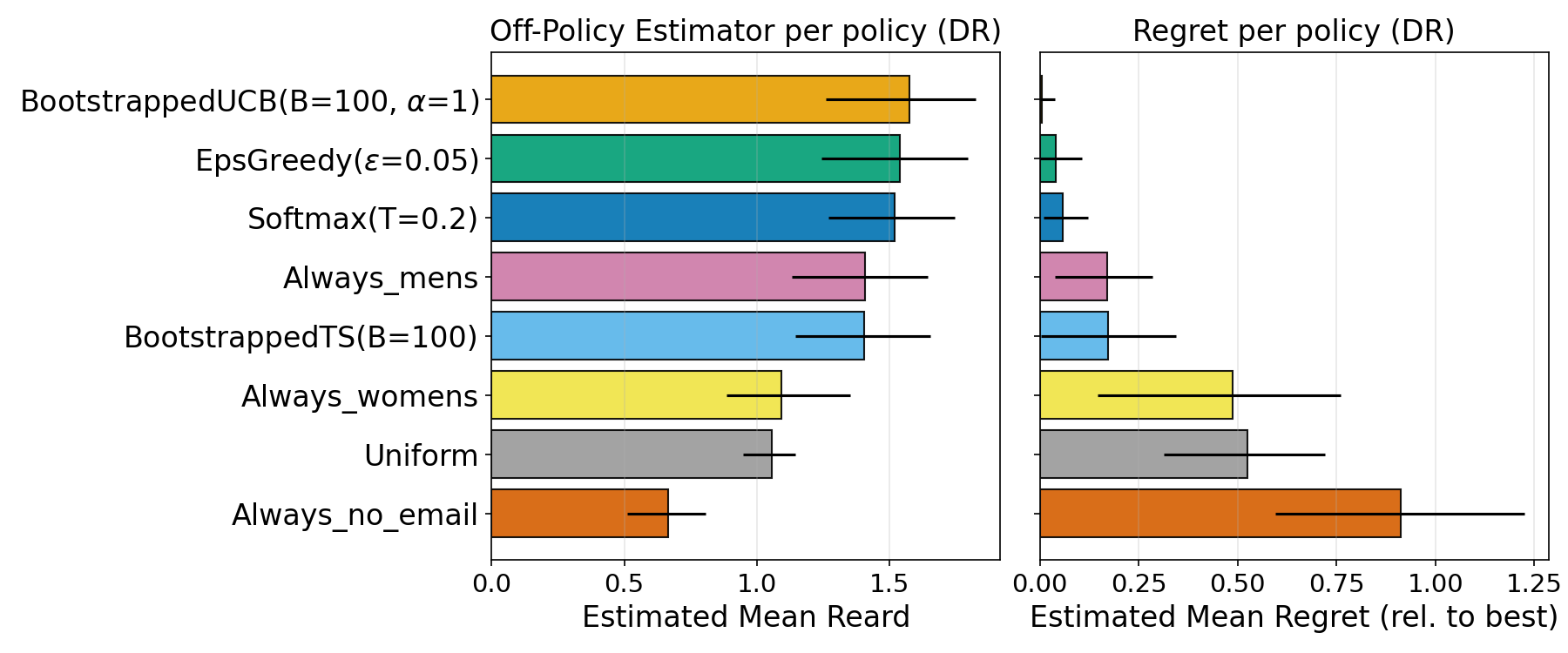} \caption{\textbf{Hillstrom Dataset} using spend as outcome.} \label{fig:hillstorm_spend} \end{subfigure}
\hfill \begin{subfigure}[b]{0.48\textwidth} \centering \includegraphics[width=\textwidth]{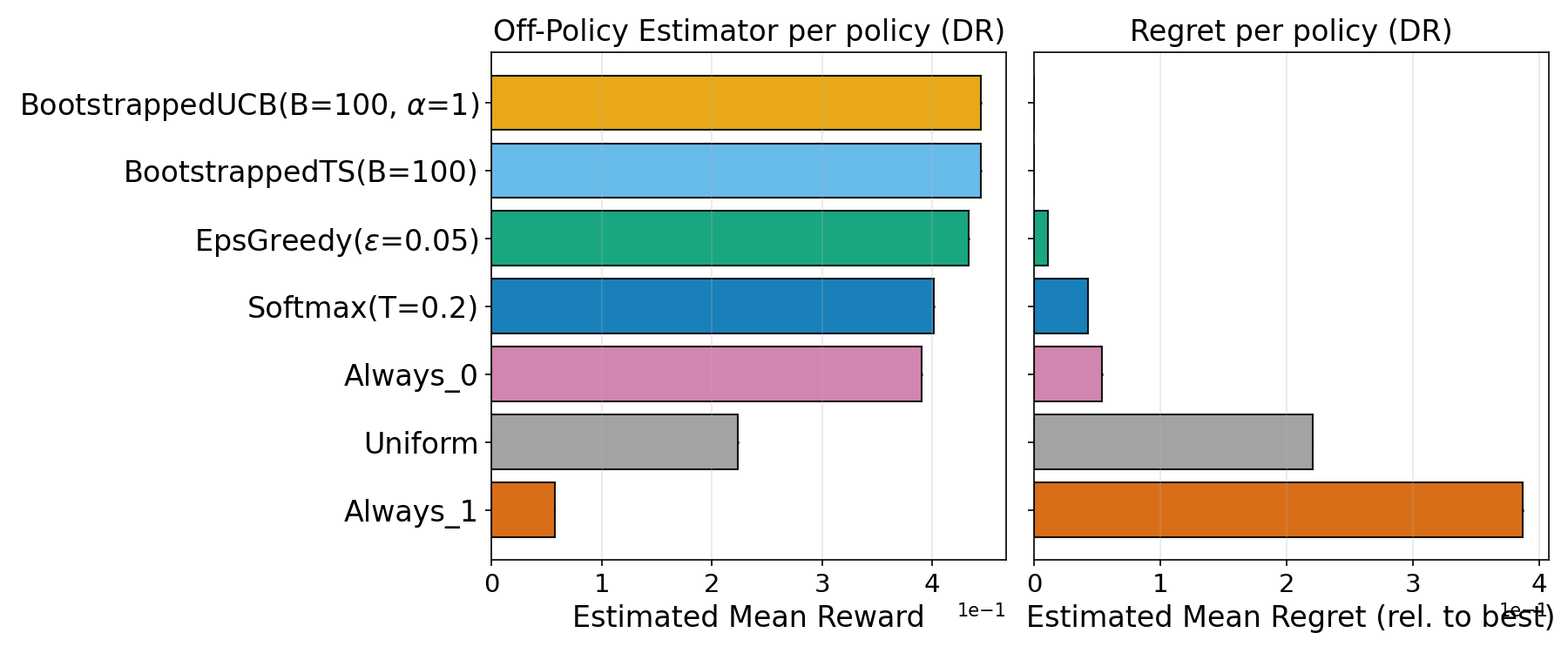} \caption{\textbf{Criteo Dataset} using conversion as outcome. } \label{fig:criteo_conversion} \end{subfigure}
\hfill\begin{subfigure}[b]{0.48\textwidth} \centering \includegraphics[width=\textwidth]{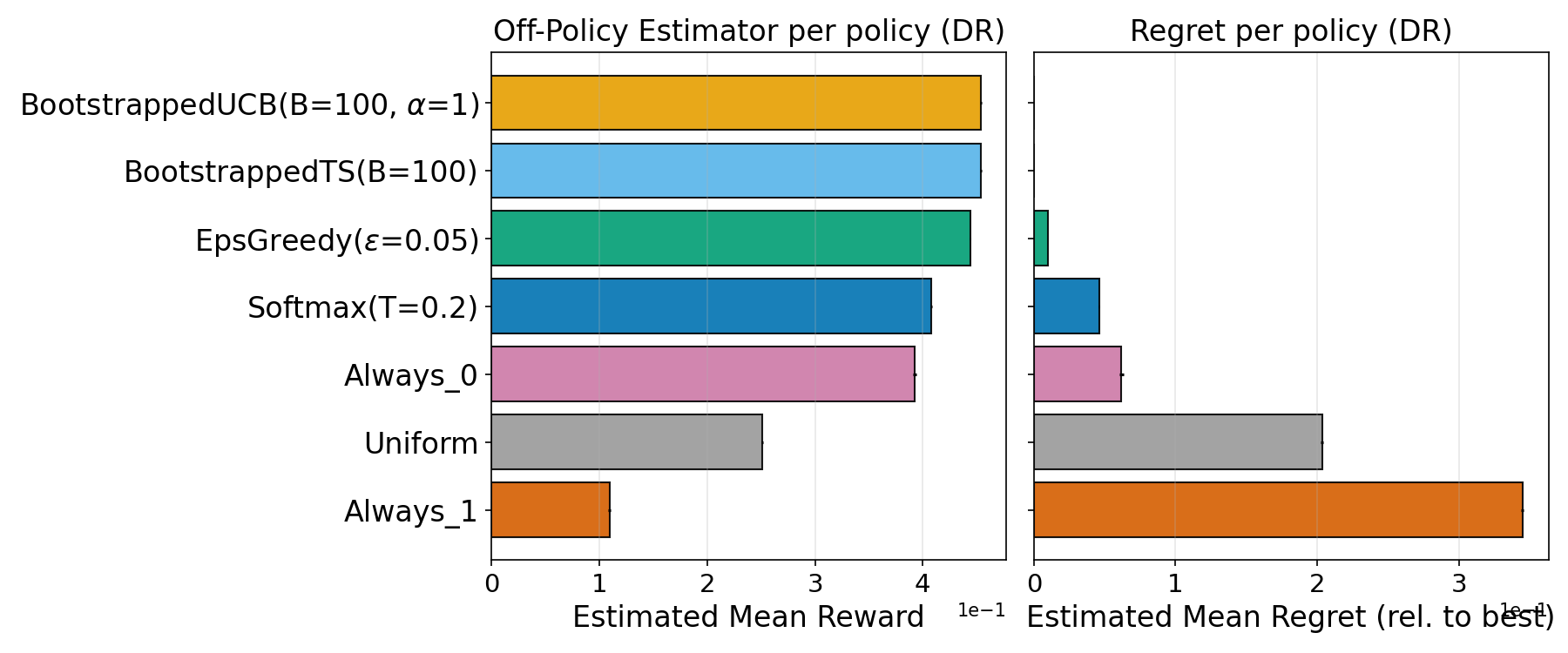} \caption{\textbf{Criteo Dataset} using visit as outcome. } \label{fig:criteo_visit} \end{subfigure}
\hfill \begin{subfigure}[b]{0.48\textwidth} \centering \includegraphics[width=\textwidth]{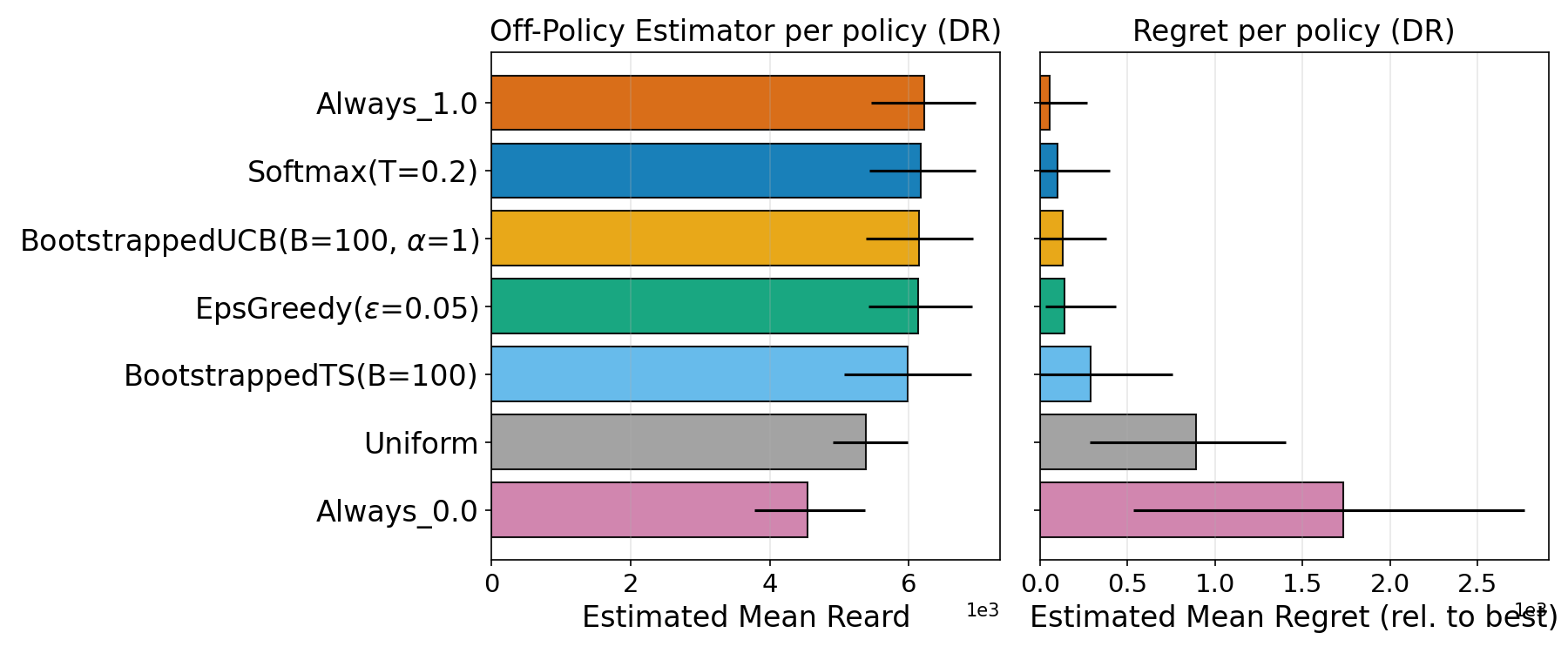}  \caption{\textbf{Lalonde Dataset.}} \label{fig:lalonde} \end{subfigure}

\caption{\textbf{OPE (DR) and regret for open Datasets.} For Hillstrom and Lalonde, adaptive policies do not clearly separate from the strongest fixed alternatives. For Criteo, adaptive policies consistently outperform non-adaptive baselines.} \label{fig:open_datasets} \end{figure}

Taken together, the open-dataset results show that OPE can serve as a practical deployment filter:
\begin{itemize}
\item \textbf{Hillstrom (Fig.\ref{fig:hillstorm_conversion}):} weak adaptivity signal. The uncertainty bands of the top adaptive policies overlap substantially with those of the best fixed alternatives, so the available evidence does not justify the added complexity of adaptive deployment.
\item \textbf{Criteo (Fig.\ref{fig:criteo_conversion}):} strong adaptivity signal. Adaptive policies achieve consistently higher estimated value and lower regret than non-adaptive baselines, indicating a setting in which contextual adaptation is likely to be worth deploying.
\item \textbf{LaLonde (Fig.\ref{fig:lalonde}):} overlap-limited regime. The best adaptive policies are not clearly separated from strong fixed alternatives, suggesting that the available support is insufficient for adaptivity to produce a decisive practical gain in this setting.
\end{itemize}
These three cases reinforce that the value of adaptivity is not universal: it depends on the strength of contextual heterogeneity and on the quality of support provided by the data.

Overall, the results support two interpretations. First, OPE successfully distinguishes settings in which contextual adaptation is worth deploying from settings in which a strong fixed policy is already sufficient. Second, when adaptive policies are beneficial, the warm-start simulation shows that several reasonable candidates already operate close to the oracle benchmark from the start of online interaction.

\section{Discussion}

Our results support a decision-oriented interpretation of adaptive experimentation. Rather than asking only which policy achieves the highest estimated value, our pipeline addresses a more practical question: whether historical A/B test data provide sufficient evidence that contextual adaptation is worth deploying at all, and, if so, whether the performance spread among strong adaptive candidates is large enough to justify deployment. Viewed through this lens, the main empirical contrast is not between the best two adaptive policies, but between adaptive and non-adaptive designs. Across synthetic settings with genuine heterogeneous treatment effects, adaptive policies consistently improve upon fixed baselines and remain close to the oracle benchmark; by contrast, in settings with weak or absent heterogeneity, the gains from adaptation are negligible. This pattern clarifies the practical role of OPE in the regimes considered here. This observation is directly relevant to the role of offline policy learning: since our objective is to compare the full policy portfolio rather than to optimize a single policy, selecting a strong candidate via OPE may already recover most of the available value, with limited marginal benefit from an additional policy-learning step.

The simulator-based online stage should be interpreted as a matched best-case benchmark for offline-to-online transfer rather than as a realistic proxy for deployment under arbitrary conditions. By design, it isolates how much of the offline-learned structure survives the transition to interaction when the data-generating process is stable and shared across stages. Our results show that OPE provides a reliable coarse deployment signal under this matched regime --- clearly for broad contrasts (adaptive versus non-adaptive designs, or strong candidates versus obviously weak baselines), though not necessarily for fine-grained ranking among history-dependent adaptive learners. The present experiments therefore do not establish robustness to offline-online mismatch; they evidence internal consistency and best-case warm-start transfer.

%A central empirical pattern in the synthetic experiments is that, whenever contextual heterogeneity is strong enough for adaptivity to improve over fixed allocation, the strongest adaptive policies already operate close to the oracle benchmark. The practically important gain therefore arises from introducing contextual adaptation in the first place, not from fine-grained optimization among the strongest adaptive candidates. This observation is directly relevant to the role of offline policy learning: since our objective is to compare the full policy portfolio rather than to optimize a single policy, selecting a strong candidate via OPE may already recover most of the available value, with limited marginal benefit from an additional policy-learning step.

The open benchmarks illustrate three distinct deployment regimes. \textit{Hillstrom} represents a relatively clean randomized benchmark with modest heterogeneity: the top adaptive policies are not clearly separated from the best fixed alternatives, so the evidence does not strongly support the additional complexity of adaptive deployment. \textit{Criteo}, by contrast, is a large-scale benchmark with meaningful covariate-driven heterogeneity, and here adaptive policies consistently achieve higher estimated value and lower regret than non-adaptive baselines; this is precisely the type of setting in which our framework would recommend adaptive experimentation. \textit{LaLonde} occupies a third regime, in which selection bias and overlap limitations complicate counterfactual comparison; here the policy-value perspective remains informative, but the results also highlight that poor support can limit the practical gains obtainable from adaptive decision rules. Taken together, these benchmarks reinforce that the value of adaptivity is not universal: it depends on the strength of contextual heterogeneity and on the quality of support provided by the data.

Our implementation also computes policy-value estimates produced by the three OPE estimators (DR, IPS, DM). Across all considered scenarios, DR either achieves the highest estimated value or is statistically indistinguishable from the best-performing estimator, as evidenced by overlapping confidence intervals. If we compare the different policies estimators for each policy in the synthetic scenarios, we have a total of 38 combinations. In the aggregate, across the 38 scenario-policy combinations considered, DR attains the largest point estimate in 8 cases, its confidence interval overlaps that of the best estimator in all 38 cases, and we observe no case in which DR is clearly dominated by a competing estimator through non-overlapping intervals. This supports the use of DR as our primary estimator and suggests that the main qualitative conclusions are robust across standard OPE choices.

While poor overlap can lead to unstable inverse-propensity weights and inflated finite-sample variance in weighting-based and doubly robust estimators, this issue is most acute when propensity scores approach the boundaries of the unit interval~\cite{li_overlap_2019,yang_overlap_2026,zhang_stable_dr_2022}. In our synthetic experiments, however, the logging propensities are known and uniformly bounded, so the heavy-weight regime that motivates these concerns does not arise. In the real-data benchmarks, we likewise do not observe instability in the bootstrap estimates or qualitative disagreement between DR, IPS, and DM, suggesting that heavy-tailed behavior does not dominate in the regimes considered here.

Our study is related to, but distinct from, the classical tradeoff between reward maximization and best-arm identification in adaptive experiments \cite{HadadEtAl2021AdaptiveExperiments}. The candidate contextual bandit policies we evaluate online are primarily reward-oriented: they are assessed through cumulative reward and regret and are therefore closer in spirit to regret minimization than to pure best-arm identification. At the same time, the synthetic experiments provide an oracle policy $\pi^\star(x)$, which allows us to inspect whether these reward-oriented policies recover the correct context-dependent decision structure. Accordingly, the plots of $\pi(x)$ versus $x$ should be interpreted as an ex post diagnostic of how closely the candidate policies approximate the oracle decision rule, not as evidence that the adaptive design itself targets a best-arm-identification objective. Our contribution is therefore not to resolve the welfare-versus-identification tradeoff within an adaptive experiment, but to use offline policy evaluation as a pre-deployment tool for deciding whether adaptive experimentation is worth pursuing and which reward-oriented policies are promising candidates.

\section{Conclusion and Future Work}

In this work, we studied how historical data from fixed randomized experiments (A/B tests) can be used to inform the deployment of adaptive experiments based on contextual bandits. Rather than optimizing a single policy from logged feedback, we focused on policy-level comparison: determining whether contextual adaptation is worth deploying and which adaptive policies are reasonable candidates given the available evidence. To this end, we combined off-policy evaluation with a controlled warm-start simulation.

Across the settings considered here, our results show that adaptive policies improve upon fixed allocation when meaningful contextual heterogeneity is present, while offering little benefit in its absence. More broadly, the paper shows that historical A/B test data can be used not only to evaluate candidate adaptive policies offline, but also to support deployment decisions and warm-start initialization before live interaction.

Our conclusions should be interpreted within the scope of the present study. First, we do not perform offline policy learning; our aim is comparative evaluation of a pre-specified policy portfolio rather than optimization over a large policy class. Second, the warm-start simulator is intentionally a matched best-case benchmark; consequently, the present results do not establish robustness to outcome-model misspecification, covariate shift, or temporal drift. Third, our setting is contextual bandits rather than full sequential reinforcement learning, so the conclusions concern one-step treatment assignment rather than long-horizon control. These scope choices are deliberate: they isolate the question the paper is meant to answer, namely whether existing A/B test data contain enough evidence to justify adaptive deployment and to select a strong adaptive policy candidate.

Several directions naturally extend this study and emerges as potential future works. First, while the present results suggest limited incremental value from offline policy learning in the regimes considered here, larger policy classes and richer action spaces may change this picture. Second, extending the warm-start framework to non-stationary or partially observable environments would bring the analysis closer to realistic deployment conditions. Third, incorporating operational constraints such as fairness, risk sensitivity, or budget limits would broaden the practical relevance of the framework. Finally, additional large-scale logged datasets or controlled live experiments would help evaluate the robustness of OPE-based deployment decisions in more complex production settings.

\subsubsection*{Acknowledgments} The authors would like to thank Daniel Vieira Batista, Marco Antonio Afonso Aragon, Adriana Laurindo Monteiro and Gabriel Mattos Langeloh for inspiring discussions that motivated the initial direction of this work.

\subsubsection*{Disclosure of Interests} Any opinions, findings, conclusions or recommendations expressed in this material are those of the authors and do not necessarily reflect the views of Itaú Unibanco and Instituto de Ciência e Tecnologia Itaú. This document is not and does not constitute or intend to constitute investment advice or any investment service. It is not and should not be deemed to be an offer to purchase or sell, or a solicitation of an offer to purchase or sell, or a recommendation to purchase or sell any securities or other financial instruments. In addition, all data used in this study comply with the Brazilian General Data Protection Law.
%
% Bibliography
\bibliographystyle{splncs04}
\bibliography{references}

\end{document}